\documentclass[preprint,12pt,authoryear]{elsarticle}

\usepackage{amssymb}
\usepackage{hyperref} % url link to github
\usepackage{amsmath} % align environment
\usepackage{mathrsfs} % Notation of nice S etc
\usepackage[utf8]{inputenc} % fuer umlaute in Bibtex
\usepackage[T1]{fontenc} % fuer umlaute in Bibtex
\usepackage{xcolor} %für farbige notizen
\usepackage{csquotes} % for enquote
\usepackage{caption} % side by side figures
\usepackage{subcaption} % side by side figures
\usepackage{cancel}
\usepackage{dsfont}

\UseRawInputEncoding
\usepackage{listings}

\newcommand{\Pcal}{\mathcal{P}}
\newcommand{\Sscr}{\mathscr{S}}
\newcommand{\Cscr}{\mathscr{C}}

\newcommand{\Scal}{\mathcal{S}}

\newtheorem{theorem}{Theorem}
\newtheorem{lemma}[theorem]{Lemma}
\newtheorem{corollary}[theorem]{Corollary}
\newdefinition{remark}{Remark}
\newdefinition{example}{Example}
\newdefinition{definition}{Definition}
\newproof{proof}{Proof}

\journal{International Journal of Approximate Reasoning}

\begin{document}

\begin{frontmatter}


\author{Hannah Blocher\corref{cor1}\fnref{label2}}
\ead{hannah.blocher@stat.uni-muenchen.de}

\author{Georg Schollmeyer\fnref{label2}}
%\ead{georg.schollmeyer@stat.uni-muenchen.de}

\author{Malte Nalenz\fnref{label2}}
%\ead{malte.nalenz@stat.uni-muenchen.de}

\author{Christoph Jansen\fnref{label2}}
%\ead{christoph.jansen@stat.uni-muenchen.de}
 \cortext[cor1]{Corresponding author}

\title{Comparing Machine Learning Algorithms by Union-Free Generic Depth}


%\author{}

 \affiliation[label2]{organization={Department of Statistics, LMU Munich},%Department and Organization
            addressline={Ludwigstr. 33}, 
            city={Munich},
            postcode={80539}, 
            state={Bavaria},
            country={Germany}}

\begin{abstract}
We propose a framework for descriptively analyzing sets of partial orders based on the concept of depth functions. 
Despite intensive studies in linear and metric spaces, there is very little discussion on depth functions for non-standard data types such as partial orders.
We introduce an adaptation of the well-known simplicial depth to the set of all partial orders, the \textit{union-free generic (ufg)} depth. 
Moreover, we utilize our ufg depth for a comparison of machine learning algorithms based on multidimensional performance measures. Concretely, we provide two examples of classifier comparisons on samples of standard benchmark data sets. Our results demonstrate promisingly the wide variety of different analysis approaches based on ufg methods. Furthermore, the examples outline that our approach differs substantially from existing benchmarking approaches, and thus adds a new perspective to the vivid debate on classifier comparison.\footnote{\label{footn: github}\textbf{Open Science:} Reproducible implementation and data analysis are available at: \url{https://github.com/hannahblo/Comparing_Algorithms_Using_UFG_Depth}}

\end{abstract}
%Graphical abstract

% mandatory
% Please use 'Highlights' in the file name and include 3 to 5 bullet points (maximum 85 characters, including spaces, per bullet point).
\begin{highlights}
\item Introducing a depth function on the set of partial orders.
\item Analyzing the empirical distribution of a sample of partial orders.
\item Developing multidimensional performance comparison.
\item Introducing a new benchmarking approach for machine learning algorithms.
\item Providing two concrete multidimensional evaluations of classification algorithms.
\end{highlights}

\begin{keyword}
partial orders \sep data depth \sep benchmarking \sep algorithm comparison \sep outlier detection \sep non-standard data
% Info from https://www.elsevier.com/journals/international-journal-of-approximate-reasoning/0888-613X/guide-for-authors
% maximum of 6 keywords, using American spelling and avoiding general and plural terms and multiple concepts




\end{keyword}

\end{frontmatter}



\section{Introduction and Related Literature}
We begin with the general motivation for this paper and an overview of the contributions of our paper to the comparison of machine learning algorithms. We also provide references to related literature.

\subsection{Motivation}
  \textit{Partial orders} -- and the systematic incomparabilities of objects encoded in them -- occur naturally in a variety of problems in a wide range of scientific disciplines. Examples range from decision theory, where the agents under consideration might be unable to arrange the consequences of their actions into total orders (see, e.g.,~\cite{sks1995,kikuti}) or have partial cardinal preferences (see, e.g.,~\cite{jsa2018,jbas2022}), over social choice theory, where a fair aggregate order might only be possible by incorporating systematic incomparabilities (see, e.g.,~\cite{p2012,jsa2018b}), to finance, where risky assets do not always have to be comparable (see, e.g.,~\cite{ll1984,c2015}). Of course, many other relevant examples exist.
  
In the specific context of statistics and machine learning, the incompleteness of the considered orders often originates from the fact that the objects to be ordered are compared with respect to several criteria and/or on several instances \textit{simultaneously}: only if there is unanimous dominance of one object over another, this order is included in the corresponding relation.
Quite a number of research papers recently have been devoted to such comparison in the specific context of classification algorithms, either with respect to multiple quality metrics~(e.g.,~\cite{ehl2012,jansen23}) or across multiple data sets (e.g.,~\cite{d2006,bcm2016}) or with respect to genuinely multidimensional performance criteria like receiver operating characteristic (ROC) curves (e.g.,~\cite{c2020}). 

Another source of partial incomparability of classifiers is the case of classifiers that make only imprecise predictions, like for example the naive credal classifier (cf., \cite{ZAFFALON20025}) or credal sum-product networks (cf., \cite{pmlr-v62-maua17a}). In this case, the imprecision in the predictions may take over to incomparabilities of the then possibly interval-valued performance measures.\footnote{For example one could think of comparing classifiers with utility-discounted predictive accuracy, cf., \cite[p. 1292 ]{ZAFFALON20121282} under the usage of a whole range $[\underline{a},\overline{a}]$ for the coefficient of risk aversion.}
 
Within the application field of machine learning and statistics, one further aspect is of special importance: Since the instances generally depend on chance, the same is true for the partial orders considered. Consequently, instead of a single partial order, random variables must then be analyzed that map into the set of \textit{all possible} partial orders on the set of objects under consideration. For example, in the aforementioned comparison of classification algorithms, the concrete order obtained depends on the random instantiation of the data set on which they are applied. In this paper, we are interested in exactly this situation: we discuss ways to \textit{descriptively} analyze samples of such partial order-valued (or short: \textit{poset-valued}\footnote{Note that in fact, we speak here about random variables that have posets (on a common underlying ground space) as outcomes. This should not be confused with random variables which have values in a partially ordered set.}) random variables.

Of course, a descriptive analysis of samples of partial orders requires a completely different mathematical apparatus than the analysis of standard data. A suitable formal framework is by no means obvious.\footnote{Many approaches proposed for analyzing poset-valued data rely on distance measures, see, e.g.,~\cite{critchlow85, brandenburg12}. For further discussion see~\cite{bsj2022}, Section~2.} Fortunately, it turns out that the concept of \textit{(data) depth function} and the data representation given by \textit{formal concept analysis} can be promisingly combined and applied to poset-valued random variables.\footnote{The representation of posets via formal concept analysis can be expressed as a closure operator, see Definition~\ref{def:closure operator posets}. Therefore, formal concept analysis is not necessary to follow this paper. For more information and discussion of this representation, see~\cite{bsj2022}.} In this paper, we adapt the concept of depth to poset-valued random variables by introducing the \textit{union-free generic (ufg) depth}. In general, depth functions define a notion of centrality and outlyingness of observations with respect to an entire data cloud or an underlying distribution, see e.g.~\cite{sz2000,m2002}. So far, depth functions have mainly been applied to $\mathbb{R}^d$-valued random variables. Exceptions are e.g. the definition of depth functions on lattices in~\cite{schollmeyer17a, schollmeyer17b}, on total orders, see~\cite{goibert22}, and a general discussion on depth functions based on formal concept analysis in~\cite{blocher23b}. Since we transfer this depth concept to poset-valued data, some classical descriptive statistics can naturally be adapted to this particular \textit{non-standard data} type.

\subsection{Comparing Machine Learning Algorithms} \label{sec:comparison}
Before turning to the main part of the paper, we first indicate which contributions (beyond benchmarking) our methodology can add to the general task of analyzing machine learning (ML) algorithms. 

While the basic task of performance comparison is very common in machine learning (cf., \cite{Hothorn} and the references therein), our methodological contribution deviates from the typical benchmark setting with regard to at least two points:

(I) First, we compare algorithms not with respect to one unidimensional criterion like, e.g., balanced accuracy, but instead look at a whole set of performance measures. 
We then judge one algorithm as at least as good as another one if it is not outperformed with respect to any of these performance measures. With this, for every data set, we get a partial order of algorithms and since we are not looking at only one, but a whole population/sample of data sets, we get a poset-valued random variable. Importantly, we do not hold the view that there is an underlying true (random) total order together with a coarsening mechanism that generates the (random) partial order. Such views are often termed \textit{epistemic}, cf., \cite{Couso_2014}. Applications of this view in the context of partial order data can be found for example in \cite{NIPS2007_fe8c15fe,nakamura2019learning}. In contrast, within the nomenclature of \cite{Couso_2014}, we see our approach more in the spirit of the \textit{ontic view} that is usually applied to set-valued data.\footnote{Concrete applications can be found e.g. in \cite{plass2015_ontic} for the ontic, and in \cite{plass2015statistical} for the epistemic view. A case where the ontic and the epistemic view coincide is discussed in \cite{pmlr-v103-schollmeyer19a}. Beyond this, in the field of partial identification in the context of generalizing confidence intervals to confidence regions for the so-called identified set, the question about an ontic vs. an epistemic view is also implicitly asked (without reference to these terms) by asking if such a confidence set should cover the true parameter or the whole identified set with prespecified probability, cf., \cite{stoye2009statistical}.} However, since in our case the random objects are partial orders, the term \textit{ontic} seems to fit not perfectly. Instead, we understand poset-valued data as a special type of non-standard data.\footnote{Note that we do not want to generally rule out epistemic treatment, but this is not the focus of this paper. Such a treatment could use that every partial order can be described by the set of all its linear extensions. For further discussions, cf.,~\cite{bsj2022}.}

(II) Second, we are not interested in which algorithm is the best or most competitive. Instead, we are interested in how the relative performance of different algorithms is distributed over a population/sample of different data sets. Analyzing the distribution of performance relations is in our view a research question of its own statistical importance that may add further insights to other types of analyses like that of \cite{jansen23,jsa2023}, which are of utmost importance when it comes to choosing between different machine learning algorithms.
%example of own interest \textcolor{red}{Hier vergleich zum Malte Paper (Georg)}

These deviations can have very different motivations: Analyzing multidimensional criteria (of performance, here) is already motivated by the fact that different performance measures might be conceptually on an equal footing (at least, if one has no further concrete, e.g., decision-theoretic desiderata at hand). Therefore it appears natural to take more than one measure at the same time into account. Beyond this, there are far more possible motivations for dealing with multidimensional criteria: For example, if one accounts for distributional shifts within covariates in classification, then for different covariate distributions the class balance of the class labels will vary. This can naturally be captured by looking at different weightings of the true positives and the true negatives within the construction of a classical performance measure.\footnote{Note that usual performance measures are more or less simple transformations of the vector of the true positives and the true negatives (and the class balance).} Alternatively, one can take into account different discrimination thresholds for the classifiers simultaneously, which would correspond to looking at a whole region of the receiver operating characteristic.

Also, the motivation for the second point can be manifold:
Generally, it seems somehow naive to search for one best algorithm per se. For example, the scope of application of an algorithm can vary very strongly and different algorithms could be best for different situations. Further, in certain situations different algorithms can be comparable in their performance with respect to one specific measure, but incomparable if one looks at different performance measures at the same time. Therefore, it can be of high interest, how the conditions between different algorithms change over the distribution of different data sets or application scenarios. For example, if in one very narrowly described data situation the performances of different algorithms vary extremely from case to case, but not so much across algorithms, then, at some point, it would become more or less hopeless to search for the best algorithm in the training phase, because one knows that in the prediction setting, the situation is too different to the training situation.

Another aspect is outlier detection: If one knows that in a large, maybe automatically generated benchmark suite there are data sets that have some bad data quality (for example if some covariates are meaningless because of some data formatting error, etc.), then it would be reasonable to try to exclude such outlying data sets from a benchmark analysis beforehand.  Candidates of such outliers are then naturally data sets with a low depth value. 

%With these motivations in mind, we now use our methodology to analyse the distribution of the poset-valued perfomance-relations between selected machine learning classification algorithms.%an example of the distributional analysis of of the multidimenional performance of classifiers.

\subsection{Outline of the Paper}
This paper is an extended version of \cite{blocher23} presented at the 13th International Symposium on Imprecise Probabilities: Theories and Applications (ISIPTA 2023) conference in Oviedo, Spain. Compared to the ISIPTA 2023 conference, this paper focuses more on the application. We added an entire section on comparison with other methods for evaluating poset-valued data and a sensitivity analysis with respect to the number of performance measures, see Section~\ref{sec:application}. We also extend the analysis and implementation of the ufg depth, see Sections~\ref{sec:properties} and~\ref{subsec:implementation}.

Our paper is organized as follows: We begin with a brief practitioner's summary, focusing on the special case of classifier comparisons. In Section~\ref{prel}, we briefly discuss the required mathematical definitions and concepts. We give a formal definition of our depth function, the ufg depth, in Section~\ref{sec:ufg} and discuss some of its properties in Section~\ref{sec:properties}. While Section~\ref{subsec:implementation} prepares our application by providing implementation details, Section~\ref{sec:application} is devoted to applying our framework to specific examples, namely the analysis of the goodness of classification algorithms on different data sets. First, we focus on comparing our benchmarking approach to further benchmarking concepts, see Section~\ref{sec: uci} and afterward, we give a detailed demonstration of the ufg method, see Section~\ref{sec: openml}. Section~\ref{sec:conclusion} concludes by elaborating on some promising perspectives for future research.
\section{Summary for Practitioners}
In this section we demonstrate how our general methodology can be used for the problem setting of selecting a reasonable classifier among  a set of candidate classifiers within an application situation.
Selection methods for classification algorithms are of immense importance for applications in a wide range of industries. Depending on the concrete branch, large amounts of money or even human lives can depend on it, e.g., treatment decision.  As many practitioners have broad expertise in their specialty but are less familiar with statistical methodology, \textit{reliable} guidance on classifier selection is crucial. To qualify as reliable, a selection method should at least meet the following standards: 
\begin{itemize}
    \item[(I)] \textbf{Option to make no decision:} If the inherent uncertainty of the comparison situation is too large to identify a clear best classifier, the method should report and not obscure this fact.
    \item[(II)] \textbf{Utilization of available information:} If the evidence is sufficient to identify a clear best classifier, the method should recognize this and not make overly cautious recommendations.
    %\item[(III)] \textcolor{orange}{\textbf{Include various performance measures:} When comparing algorithms, the question soon arises as to which measures should be used to evaluate the performance of the algorithms. The method should provide a clear and easily accessible explanation of how to handle the different performance measures.}
    \item[(III)] \textbf{Intuitive accessibility:} The method should enable the user to develop an intuitive understanding of it (an exact understanding of the underlying mathematics is not necessary).
\end{itemize}
We present a framework for comparing classifiers that addresses requirements (I) to (III) and is based on two simple observations: \textit{First}, the suitability of a classifier depends on the specific question, i.e., the specific data set being analyzed. Accordingly, the search should not (solely) be for the best classifier per se, but represent how \textit{central/typical} a data set and the induced classifier ranking is. \textit{Second}, the performance of certain classifiers might be incomparable already on one data set because, for example, multiple or vaguely defined performance measures are considered that conflict each other in the quality dimensions.

These two simple observations serve as a guiding principle of our construction: To compare data sets in terms of their centrality (think of the empirical \textit{median} as most central one and the high and low \textit{quantiles} as more outlying ones), we use the statistical theory of \textit{data depth}: the deeper a data set lies (in some abstract space), the more central it is. To account for the complexity of the (potentially multidimensional or vaguely defined) concept of performance, we allow classifiers to be only \textit{partially} ordered (already when considering only a single data set).

Our resulting framework does indeed manage to meet the above standards: By using partial orders rather than total ones, in the spirit of (I), a clear best classifier is identified only if the underlying evidence is strong enough and always accompanied by information on the typicality of its superiority. In the spirit of (II), our framework avoids the incomparability of classifiers if the evidence is strong enough for a clear favorite (who is still accompanied by information on the typicality of its superiority). Finally, standard (III) is met as well: Since data depth is a generalization of the elementary statistical concept of a \textit{quantile} (where, e.g., the deepest point corresponds to the median), every user is directly guided by natural intuition when applying the proposed concepts to real-world applications.

In a nutshell, our contributions can be summarized as follows: We offer practitioners the opportunity to reliably compare classification algorithms with regard to several performance measures and taking into account the typicality/centrality of the specific situation analyzed. Beyond the theoretical soundness of our framework, we provide well-documented implementations of our method, which are easily transferable to comparable applications. Last but not least, we propose a customized depth function for our application, the \textit{ufg depth}, and enable the reliable comparison of classifiers by performing the following simple three-step scheme:
\begin{enumerate}
    \item Compute the ufg depth function. We provide an R-script that can be easily adapted. The script can be found on GitHub, see \url{https://github.com/hannahblo/Comparing_Algorithms_Using_UFG_Depth}.
    \item Use the partial order among classifiers corresponding to the highest depth value as a reference and the undominated algorithms as recommendations. We perform this analysis in Section~\ref{sec: openml} on Figure~\ref{fig:max_min}.
    \item Check how stable the dominance structure is by checking what the partial orders with the $k$ highest depth values have in common. An example analysis using the interesection of the deepest posets can be found in the Section~\ref{sec: openml} discussing Figure~\ref{fig:Schnitt}.
\end{enumerate}
Finally, we want to emphasize that our concept can be used to compare any objects that arise from a sample of instances and where the comparison is based on multiple numeric measures. Thus, the concept introduced here is not limited to classification problems.

\section{Preliminaries} \label{prel}
\textit{Partial orders (posets)} sort the elements of a set $M$, where we allow that two elements $y_1, y_2 \in M$ are incomparable. Formally stated: Let $M$ be a fixed set. Then $p \subseteq M \times M$ defines a partial order (poset) on $M$ if and only if $p$ is \textit{reflexive} (for each $y \in M$ holds $(y, y)\in p$), \textit{antisymmetric} (if $(y_1, y_2), (y_2,y_1) \in p$ then $y_1 = y_2$ is true) and \textit{transitive} (if $(y_1, y_2), (y_2,y_3) \in p$ then also $(y_1,y_3) \in p$). If $p$ is also strongly connected (for all $y_1,y_2 \in M$ either $(y_1,y_2) \in p$ or $(y_2, y_1) \in p$), then $p$ defines a \textit{total/linear order}. The reverse, where the poset consists only of the reflexive part, is called \textit{trivial order} $p_{\Delta}$.
For a fixed set $M$, various posets sort the set $M$. In this paper, we are interested in all posets that can occur on the set $M$ with cardinality $\# M$ being finite. We denote the set of all posets on $M$ by $\Pcal_M$ (or $\Pcal$ for short). Sometimes it can be useful to consider only the \textit{transitive reduction} of a poset $p$, this means that for poset $p$ we delete all pairs $(y_1, y_2)$ which can be obtained by a transitive composition of other elements in $p$. We denote the transitive reduction of a poset $p$ by $tr(p)$. Note that there exists a one-to-one correspondence between the transitive reduction of posets and the posets itself. In particular, this transitive reduction is often used to simplify the diagram used to represent the partial order. These diagrams are called \textit{Hasse diagrams}. They consist of edges and knots where the knots represent the elements of $M$ and the edges state the relation between the elements. More precisely, if $(a,b) \in p$ for a poset $p$, then there is a path from $a$ to $b$ that points strictly upwards, e.g., see Figure~\ref{fig:max_min}. The reverse, where we add all pairs that follow from transitivity, is called \textit{transitive hull}. We denote by $th(p)$ the transitive hull of an antisymmetric and reflexive relation $p$. We refer to \cite{ganter12} for further readings on partial orders.
From now on, let $\Pcal$ be all posets for a fixed set $M$. We denote the elements of $M$ by $y$. 

The concept for analyzing poset-valued observations presented in this paper is based on a \textit{closure operator} on $\Pcal$, see Section \ref{sec:ufg}. In general, a closure operator $\gamma_{\Omega}: 2^{\Omega} \to 2^{\Omega}$ on set ${\Omega}$ is an operator which is \textit{extensive} (for $A \subseteq {\Omega}, \: A \subseteq \gamma_{\Omega}(A)$ holds), \textit{isotone}, (for $A,B \subseteq {\Omega}$ with $A\subseteq B,\: \gamma_{\Omega}(A) \subseteq \gamma_{\Omega}(B)$ is true) and \textit{idempotent} (for $A \subseteq {\Omega}, \; \gamma_{\Omega}(A) = \gamma_{\Omega}(\gamma_{\Omega}(A))$ holds). The set $\gamma_{\Omega}(2^{\Omega})$ is called the \textit{closure system} and we say that $\gamma(B)$ for $B \subseteq \Omega$ is the \textit{closure} of $B$. In what follows, we use that every closure operator (and therefore the closure system) can be uniquely described by an implicational system. Generally, an \textit{implicational system} $\mathcal{I}$ on ${\Omega}$ is defined by a subset of $2^{\Omega} \times 2^{\Omega}$. The implicational system corresponding to the closure operator $\gamma_{\Omega}$ is defined by all pairs $(A,B) \in 2^{\Omega} \times 2^{\Omega}$ satisfying $\gamma_{\Omega}(A) \supseteq \gamma_{\Omega}(B)$. For short, we denote this by $A \to B$. For more details on closure operators and implicational systems, see \cite{bertet18}.

We aim to define a centrality and outlyingness measure on the set of all posets $\Pcal$ based on a fixed and finite set $M$. In general, functions that measure the centrality of a point with respect to an entire data cloud or an underlying distribution are called \textit{(data) depth functions}. Depth functions on $\mathbb{R}^d$ have been studied intensively by, e.g., \cite{sz2000} and \cite{mosler22}, and various notions of depth have been defined, such as Tukey's depth, see \cite{tukey75}, and simplicial depth, see~\cite{liu90}. The idea behind the ufg depth introduced here is an adaptation of the \textit{simplicial depth} on $\mathbb{R}^d$ to posets. The simplicial depth on $\mathbb{R}^d$ is based on the convex hull/closure operator that is defined as follows:
	\begin{align*}
		\gamma_{\mathbb{R}^d}\colon \begin{array}{l}
		2^{\mathbb{R}^d} \to 2^{\mathbb{R}^d}\\
		A \mapsto \left\{x \in \mathbb{R}^d \bigg| \begin{array}{l}x = \sum_{i=1}^k \lambda_i a_i \text{ with }  a_i \in A, \\ \lambda_i \in [0,1], \sum_{i=1}^k \lambda_i = 1, k \in \mathbb{N}\end{array} \right\}.
	\end{array}
	\end{align*}
	The simplicial depth considers those sets $\gamma(A)$ with $A \subseteq \mathbb{R}^d$ where the cardinality of $A$ is $d+1$ and $A$ forms a $(d+1)$ simplex (when no duplicates occur). Before going on, let us take a closer look at the set of all $(d+1)$ simplexes.  First, the set of all $(d+1)$ simplexes is a proper subset of $2^{\mathbb{R}^d}$. By using Carath{\'e}odorys' Theorem on convex sets, see \cite{eckhoff93}, we obtain that any set $B$ of $d+1$ unique points is the smallest set such that no family of proper subsets $(A_i)_{i \in \{1, \ldots, \ell\}}$ with $A_i \subsetneq B$ and $\bigcup_{i \in \{1, \ldots, \ell\}}\gamma_{\mathbb{R}^d}(A_i) = \gamma_{\mathbb{R}^d}(B)$ exists. Moreover, the closure of every set $B$ which consists of more than $d+1$ points can be divided in the sense that there exists a family of proper subsets $(A_i)_{i \in \{1, \ldots, \ell\}}$ with $A_i \subsetneq B$ such that $\bigcup_{i \in \{1, \ldots, \ell\}}\gamma_{\mathbb{R}^d}(A_i) = \gamma_{\mathbb{R}^d}(B)$. Thus, the $(d+1)$ simplexes still completely characterize the corresponding closure system. Based on this set, the simplicial depth of a point $x \in \mathbb{R}^d$ is the probability that $x$ lies in $\gamma(A)$ where $A$ consists of $d+1$ many points that are randomly drawn from the underlying (empirical) distribution.
 %every element $B \in \gamma_{\mathbb{R^d}}(2^{\mathbb{R}^d})$ can be given by a union of the convex closure of $(d+1)$-simplices. 
 For $\mathcal{M}$ being the set of all probability measures on $\mathbb{R}^d$ the simplicial depth is then given by
    		\begin{align*}
    			D\colon  \begin{array}{l} \mathbb{R}^d \times \mathcal{M} \to [0, 1],\\ (x,\nu) \mapsto \nu(x \in \gamma_{\mathbb{R}^d}\{X_1, \ldots, X_{d+1}\}), \end{array}
    		\end{align*}
    		where $X_1, \ldots, X_{d+1} \overset{i.i.d.}{\sim} \nu$ with i.i.d. being \textit{independent and identically distributed} for short. When we consider a sample $x_1, \ldots, x_n \in \mathbb{R}^d; \: n \in \mathbb{N}$, we use the empirical probability measure instead of a probability measure $\nu$. Thus, for a sample $x_1, \ldots, x_n \in \mathbb{R}^d$ with empirical measure $\nu_n$ we obtain as empirical simplicial depth
    		\begin{align*}
    			D_n\colon \begin{array}{l}
    			\mathbb{R}^d \to [0,1], \\
    			x \mapsto\binom{n}{d+1} \sum_{1 \le i_1 < \ldots < i_{d+1} \le n} \mathds{1}_{\gamma_{\mathbb{R}^d}\{x_{i_1}, \ldots, x_{i_{d+1}}\}}(x).
    			\end{array}
    		\end{align*}
    		Hence, if $x_1, \ldots, x_n$ are affine independent, then the depth of a point $x$ is the proportion of $(d+1)$ simplexes given by $x_1, \ldots, x_n$ that contain $x$.

\section{Union-Free Generic Depth on Posets} \label{sec:ufg}
Now, we introduce the union-free generic (ufg) depth function for posets which is in the spirit of the simplicial depth function, see Section~\ref{prel}. To define the depth function, we start, similar to the simplicial depth, by defining a closure operator on $\Pcal$. This closure operator maps a set of posets $P$ onto the set of posets where each poset is a superset of the intersection of $P$ and a subset of the union of $P$. In other words, any $p$ lying in the closure of $\gamma(P)$ satisfies the following condition: First, every pair $(y_1, y_2) \in M \times M$ that lies in every poset in $P$ is also contained in $p$, and second, for every pair $(y_1, y_2)$ that lies in $p$, there exists at least one $\tilde{p} \in P$ such that $(y_1, y_2)\in\tilde{p}$.
Note that while the intersection of posets defines a poset again, this does not hold for the union. 
%Therefore, we use the formal context introduced in Section~\ref{prel} and we get the closure operator by composing the two derivation operators, compare to \cite{blocher22}. With this, we get
\begin{definition}\label{def:closure operator posets}
Let $\Pcal$ be the set of posets on $M$. We define the mapping
\begin{align*}
	\gamma\colon \begin{array}{l}
		2^{\Pcal} \to 2^{\Pcal}\\
		P \mapsto \left\{p \in \Pcal \mid \bigcap\limits_{
  \tilde{p}\in P}\tilde{p} \subseteq p \subseteq \bigcup\limits_{\tilde{p} \in P}\tilde{p} \right\}.
	\end{array}	
\end{align*}
\end{definition}

\begin{remark}
    This definition is based on the theory of formal concept analysis, see \cite{ganter12}, and the formal context introduced in~\cite{bsj2022}.  Using this formal context and the corresponding theory, we immediately obtain that $\gamma$ defines a closure operator on $\Pcal$ with associated closure system $\gamma(2^{\Pcal})$.
\end{remark}

Analogously to the definition of the simplicial depth, we now only consider a subset of $2^{\Pcal}$. We denote this family by $\Sscr$.
$\Sscr$ is a proper subset of $2^\Pcal$, see Theorem \ref{th: S+card 2 and connect} for details, which reduces $2^\Pcal$ by redundant elements in the following sense: First, all subsets $P\subseteq \Pcal$ with $\gamma(P) = P$ are trivial and therefore not included in $\Sscr$. Second, if there exists a proper subset $\tilde{P} \subsetneq P$ with $\gamma(\tilde{P}) = \gamma(P)$, then $P$ is also not in $\Sscr$. %This follows by setting $\ell =1$ and $A_1 = \tilde{P}$, which defines a family contradicting Condition (C2). 
These two properties can be generalized to arbitrary closure systems, and referring to \cite{bastide00}, we call a set fulfilling these two properties \textit{generic}. The final reduction is to delete also all sets $P$ where $P$ can be decomposed by a family of proper subsets $(A_i)_{i \in \{1, \ldots, \ell\}}$ of $P$. Moreover, the union of $(\gamma(A_i))_{i \in \{1, \ldots, \ell\}}$ needs to equal $\gamma(P)$. Note that due to isotonicity, the assumption $\cup_{i \in \{1, \ldots, \ell\}}\gamma(A_i) \subseteq \gamma(P)$ is always true. We call sets respecting this third part \textit{union-free}. Thus $\Sscr$ consists of elements that are union-free and generic. The following definition summarizes these conditions.
\begin{definition}
Let $\Pcal$ be the set of posets on the finite set $M$. We set
$$\Sscr = \left\{P \subseteq\Pcal \mid \text{Condition } (C1) \text{ and } (C2) \text{ hold for } P \right\}$$ 
with Conditions (C1) and (C2) given by:
\begin{enumerate}
    \item[(C1)] $P \subsetneq \gamma(P),$
    \item[(C2)] There does not exist a family $(A_i)_{i \in \{1, \ldots, \ell\}}$ such that for all $i \in \{1, \ldots, \ell\}$ $A_i\subsetneq P$ and $\bigcup_{i \in \{1, \ldots, \ell\}} \gamma(A_i) = \gamma(P).$\footnote{In formal concept analysis this is sometimes called \textit{proper}, see~\cite[p. 81]{ganter12}.}
\end{enumerate}
We call $\Sscr$ the family of union-free generic sets.
\end{definition}

\begin{example}\label{example}
    As a concrete example, consider the set $\Sscr$ based on all posets on $\{y_1, y_2, y_3\}$. Let $p_1, p_2$ and $p_3$ be posets given by the transitive hull of $\{(y_1, y_2)\}$, $\{(y_1, y_2), (y_1, y_3)\}$ and $\{(y_1, y_3), (y_2, y_3)\}$. One can show that $\gamma(\{p_1, p_3\}) =  \gamma(\{p_1, p_2, p_3\})$. Thus, $\{p_1, p_2, p_3\}$ contradicts Condition (C2). For a single poset $p$ we can immediately prove that the closure contains only itself. Therefore, any set consisting of only one poset does not satisfy Condition (C1). In contrast, $\{p_1, p_3\}$ satisfies both Condition (C1) and Condition (C2), since it implies the trivial poset $p_\Delta:=\{(y,y)\mid y \in M\}$. Thus, $\{p_1, p_3\}$ is an element of $\Sscr$.
\end{example}

%Applying such a reduction on the set $2^{\mathbb{R}^d}$ by use of the convex closure operator, see Section \ref{prel}, we obtain all subsets $B\subseteq \mathbb{R}^d$ with cardinality smaller or equal to $d+1$.

Now, we define the \textit{union-free generic (ufg) depth} of a poset $p$ to be the weighted probability that $p$ lies in a randomly drawn element of $\Sscr$.
\begin{definition}\label{def: population ufg depth}
Let $\mathcal{M}$ be the set of probability measures on $\Pcal$ equipped with $2^{\Pcal}$ as $\sigma$-field. The \textit{union-free generic (ufg for short) depth on posets} is given by 
\begin{align*}
	D\colon \begin{array}{l}
	 \Pcal \times \mathcal{M} \to [0,1] \\
	 (p, \nu) \mapsto \begin{cases}
            0, \qquad \text{if for all } S\in \Sscr\colon \prod_{\tilde{p} \in S}\nu(\{\tilde{p}\}) = 0\\
	      c\sum_{S \in \Sscr \colon p \in \gamma(S)} \prod_{\tilde{p} \in S}\nu\left(\{\tilde{p}\}\right),\qquad\text{else} 
  \end{cases}
\end{array}		
\end{align*}
 with $c = \left(\sum_{S \in \Sscr} \prod_{\tilde{p} \in S}\nu_n\left(\{\tilde{p}\}\right)\right)^{-1}$.\footnote{Note that Condition (C1) and (C2) can be applied to the convex closure operator on $\mathbb{R}^d$, see Section \ref{prel}, and we obtain an adapted $\Sscr_{convex}$. Then, $\Sscr_{convex}$ together with  $\mathcal{M}_{convex}$ the set of measures that are absolute continuous to the Lebesgue measure, leads to the simplicial depth.}
 \end{definition}
 The two cases in Definition~\ref{def: population ufg depth} are needed because the constant $c$ is not defined in the first case. Note that if there exists an $S \in \Sscr$ with $\prod_{\tilde{p} \in \Sscr}\nu(\tilde{p}) \neq 0$, then $D \not\equiv 0$. The case that $D \equiv 0$ only occurs in two specific situations that result from the structure of the probability mass, see the non-triviality property in Corollary~\ref{cor: Dn always zero} in Section~\ref{sec:properties} for details. Note that in contrast to the simplicial depth where only sets of cardinality $d+1$ are considered, the elements of $\Sscr$ differ in their cardinality. Thus, different approaches on how to include the different cardinalities are possible, i.e., by weighting. In Definition~\ref{def: population ufg depth} we use weights equal to one. 
 
 The empirical version of the ufg depth uses the empirical probability measure $\nu_n$ given by a sample of posets $\underline{p} = (p_1, \ldots, p_n), \: n \in \mathbb{N}$ instead of the probability measure $\nu$ in Definition~\ref{def: population ufg depth}. Thus, the empirical ufg depth of a poset $p$ is therefore the normalized weighted sum of drawn sets $S \in \Sscr$ which imply $p$. 

\begin{definition}\label{def: empirical ufg depth}
 Let $\nu_n$ be an empirical probability measure based on sample $\underline{p} = (p_1, \ldots, p_n), \: n \in \mathbb{N}$ (equipped with $2^{\Pcal}$ as $\sigma$-field). The \textit{empirical union-free generic (ufg) depth} is then given by
\begin{align*}
	D_n \colon \begin{array}{l}
		\Pcal \to [0,1] \\
		p \mapsto  \begin{cases}
            0, \qquad \text{if for all } S\in \Sscr\colon \prod_{\tilde{p} \in S}\nu_n(\{\tilde{p}\}) = 0\\
            c_n \sum_{S \in \Sscr, p \in \gamma(S)} \prod_{\tilde{p} \in S}\nu_n\left(\{\tilde{p}\}\right), \qquad \text{else}
              \end{cases}
	\end{array}
\end{align*}
 with $c_n = \left(\sum_{S \in \Sscr} \prod_{\tilde{p} \in S}\nu_n\left(\{\tilde{p}\}\right)\right)^{-1}$.
\end{definition}
Note that when restricting $\Sscr$ to the set $\Sscr_{obs}:= \{ S \in \Sscr \mid \mbox{ all } p \in S \mbox{ are observed} \}$, this does not change the depth value. This holds since for other elements $S \in \Sscr$, the empirical measure for at least one $p \in S$ is zero.
 
\begin{example}\label{example_2}
    Returning to Example~\ref{example}, suppose that we observe $(p_1, p_2, p_3)$. Then for the trivial poset $p_\Delta$, the empirical ufg depth is $D_n(p_\Delta) = 1/2$. For the set $p_4$ given by the transitive hull of $\{(y_3, y_1)\}$, the value of the empirical ufg depth is zero.
    For $p_{total}$ given by the transitive hull of $\{(y_1, y_3), (y_3, y_2)\}$, the empirical ufg depth value is again zero.
\end{example}

\section{Properties of the UFG Depth and $\Sscr$}\label{sec:properties}
For a better understanding of the ufg depth, we now discuss some properties of $D$, $D_n$, and $\Sscr$. The properties of $D_n$ and $D$ describe the mutual influence between the (empirical) measure and the ufg depth.

\subsection{Properties of $\Sscr$}
We begin with introducing some properties of $\Sscr$. Later, these properties are used to analyze the (empirical) ufg depth and to improve the computation. 
First, we want to introduce a second equivalent definition of Condition~(C1) and Condition~(C2). This says that $S \in \Sscr$ if and only if there exists a poset $q$ such that every poset $p \in S$ contributes to $q$. More precisely, for every $p\in S$ we have that either $p$ has an edge $(y_1, y_2)$ which only $p$ and $q$ share, or $p$ is the only poset in $S$ which does not have an edge $(y_1, y_2)$ which also $q$ does not have. Note that if we have ensured that every poset $p \in S$ contributes to a poset $q \in \gamma(S)$, then $S$ satisfies Conditions~(C1) and~(C2). 
This will help later to prove the connectedness property, see Theorem~\ref{th: S+card 2 and connect}, give an upper bound for $\# S$ with $S \in \Sscr$, see Theorem~\ref{tighterbound}, and to improve the implementation, see Section~\ref{subsec:implementation}.
\begin{lemma}\label{lem:S_andere_def}
    Let $S \subseteq \mathcal{P}$. Then $S \in \Sscr$ if and only if there exists a poset $q \in \gamma(S) \setminus S$ such that for all $p \in S, \: q \not\in \gamma(S\setminus \{p\})$.
\end{lemma}
\begin{proof}
    Let us first assume that $S \in \Sscr$. Then, due to Condition (C1), we know that $\gamma(S) \setminus S$ is nonempty. Since $\left( (S\setminus \{x\})_{x \in S} \right)$ is a family of proper subsets of $S$, we obtain by Condition~(C2) that there must exist an element $q \in \gamma(S) \setminus S$ such that for all $x \in S, \: q \not\in \gamma(S\setminus \{x\})$.

    Conversely, suppose that there exists $q \in \gamma(S) \setminus S$ such that for all $x \in S, \: q \not \in \gamma(S\setminus \{x\})$. Condition~(C1) follows immediately. To prove Condition~(C2), let $(A_i)_{i \in \{1, \ldots, \ell\}}$ be a family of all sets with $\: A_i\subsetneq S$ for all $i \in \{1, \ldots, \ell\}$. Then for every $i \in \{1, \ldots, \ell\}$ there exists an $p_i \in S$ with $p_i \not\in A_i$ (follows from $A_i \subsetneq S$). Since $\gamma$ is isotone, we know that $\gamma(A_i) \subseteq \gamma(S \setminus\{p_i\})$. By assumption we get that $q \not\in \gamma(S \setminus\{x_i\})$ and, thus, $q \not\in \gamma(A_i)$. This argument holds for every $i\in\{1, \ldots, \ell\}$ and we obtain $q \not\in \bigcup_{i \in \{1, \ldots, \ell\}} \gamma(A_i)$. With $(A_i)_{i \in \{1, \ldots, \ell\}}$ arbitrarily chosen, the claim follows.
\end{proof}

\begin{definition}\label{def: ufg element}
    We call such a poset $q$ given by Lemma~\ref{lem:S_andere_def} an \textit{ufg element} w.r.t. $S$.
\end{definition}

As we pointed out above, if $q$ is an ufg element w.r.t. some set $S \in \Sscr$, then each poset $p \in S$ has two different ways of contributing to $q$: Either it has an edge that only $p$ and $q$ share, or it has an edge that only $p$ and $q$ do not have.
%Based on Lemma~\ref{lem:S_andere_def} and the definition of the closure operator on posets, see Definition~\ref{def:closure operator posets}
With this we immediately get a description of the ufg elements w.r.t. some set $S \in \Sscr$. 
Therefore we define the following two sets, which sort the posets into two sets w.r.t. $S \in \Sscr$. One consists of those posets that have an edge that all other posets in $S$ do not have, and conversely the other set contains all posets that do not have an edge that all other posets in $S$ have. Note that these two sets are not necessarily disjoint.
\begin{definition}\label{def:disting}
Let $S \subseteq \mathcal{P}$ and $p \in S$. We define
\begin{align*}
	&D_{S}^{p, \text{edge}} = \{(a,b)\in M\times M \mid (a,b) \in p \text{ and } \forall \tilde{p} \in S \setminus \{p\} \text{ we have } (a,b) \not\in \tilde{p} \},\\
	&D_{S}^{p, \cancel{\text{edge}}} = \{(a,b) \in M \times M \mid (a,b) \not\in p \text{ and } \forall \tilde{p} \in S \setminus \{p\} \text{ we have } (a,b) \in \tilde{p}\} .
\end{align*}
\end{definition}

%\begin{remark}
%Let $S\subseteq \mathcal{P}$ be a set such that there exists a poset $q \in \gamma(S)$ with for all $p \in S$ we have $q \not\in \gamma(S\setminus \{p\})$. Then this is equivalent to the condition that for each poset $p \in S$ there exists at least one edge $(y_1, y_2) \in p$ (or $(y_1, y_2) \not\in p$) which is given (or not given) by every other $\tilde{p} \in S \setminus \{p\}$. This follows immediately by Lemma~\ref{lem:S_andere_def} and the definition of the closure operator. More precisely, for every poset $p \in S$ we have the following two cases:
%\begin{enumerate}
%	\item There exists an edge $(y_1, y_2) \in p \cap q$ which no other poset $\tilde{p} \in S\setminus \{p\}$ has. Thus, to ensure that $q \subseteq \cup_{p \in S} p$ is true, $p$ is needed.
%	\item There exists an edge $(y_1, y_2) \in (M \times M) \setminus p \cap  (M \times M) \setminus q$ which every other poset $\tilde{p} \in S\setminus \{p\}$ has. Thus, to get $q \supseteq \cap_{p \in S} p$, poset $p$ is needed,
%\end{enumerate}
%\end{remark}

 $D_{S}^{p, \text{edge}}$ consists of all edges $(a,b)$ which the poset $p$ has, but which each other poset $\tilde{p} \in S \setminus\{p\}$ does not have. Reverse, $D_{S}^{p, \cancel{\text{edge}}}$ consist of all edges which the poset $p$ does not have, but every other poset $\tilde{p} \in S\setminus\{p\}$ does have.

Using these definitions we obtain by Lemma~\ref{lem:S_andere_def} and its corresponding discussion a concrete definition of an ufg element w.r.t. $S$ if $S \in \Sscr$, see the next corollary.
\begin{corollary}\label{cor:ufg elements}
Let $S \subseteq \mathcal{P}$. Then $q$ is an ufg element w.r.t. $S$ if and only if for all $p \in S$ we have that $q \cap D_{S}^{p, \text{edge}} \neq \emptyset$ or $(M \times M) \setminus q \cap D_{S}^{p, \cancel{\text{edge}}} \neq \emptyset$.

Moreover, for every ufg element $q$ w.r.t. $S$ there exists an ufg element $\tilde{q} \subseteq q$ w.r.t. $S$ given by
\begin{align}\label{eq: ufg minimal element}
	\tilde{q} = th\left(  \left(\bigcap_{p \in S} p \right) \cup \left( \bigcup_{p \in \tilde{S}}(y^p_1, y^p_2) \right)\right).
\end{align}
with $\tilde{S} = \{p \in S\mid D_{S}^{p, \cancel{\text{edge}}} = \emptyset\}$ and for all $p \in \tilde{\Scal}$ we set $(y_1^p, y_2^p)$ to be precisely one edge $(y_1^p, y_2^p) \in D_{S}^{p, \text{edge}}$. 
\end{corollary}
\begin{proof}
    The first part follows directly from the discussion corresponding to Definition~\ref{def: ufg element} and Lemma~\ref{lem:S_andere_def} . Therefore, we focus on showing that for each ufg element $q$ w.r.t. $S \in \Sscr$ there exists a further ufg element $\tilde{q} \subseteq q$ that is given by Equation~(\ref{eq: ufg minimal element}). For each $p \in S$ we know that $q \cap D_{S}^{p, \text{edge}} \neq \emptyset$ or $(M \times M) \setminus q \cap D_{S}^{p, \cancel{\text{edge}}} \neq \emptyset$ is true. In particular by the definition of $\tilde{S}$, we obtain that for all $\tilde{p} \in \tilde{S}$ we know $D_{S}^{\tilde{p}, \text{edge}} \neq \emptyset$. Thus, we get  $(y_1^{\tilde{p}}, y_2^{\tilde{p}}) \in D_{S}^{\tilde{p}, \text{edge}}$ for all $\tilde{p} \in \tilde{S}$. We now use these edges $(y_1^{\tilde{p}}, y_2^{\tilde{p}})$ to define $\tilde{q}$ as in Equation~(\ref{eq: ufg minimal element}). With $q\in \gamma(S)$ we know that $\cap_{p \in S} \: p \subseteq q$ needs to be true and therefore, we obtain $th(\tilde{q}) \subseteq th(q) = q \subseteq \cup_{p \in S} \: p$ which proves the claim.
\end{proof}

Note that there are cases where $q$ and $\tilde{q}$ of Corollary~\ref{cor:ufg elements} are different. Recall Example~\ref{example}. Then $\{p_2, p_3 \} \in \Sscr$ and $q = \{(y_1, y_2)\} \in \gamma(\{p_2, p_3\})$ is an ufg element with respect to $\{p_2, p_3\}$. Furthermore, $\tilde{q} = p_{\Delta}$ is also an ufg element w.r.t. $\{p_2, p_3\}$, which can be given by Equation~(\ref{eq: ufg minimal element}). We have $\tilde{q} \subseteq q$.

After discussing the different definitions of sets in $\Sscr$, let us now consider the claims made in Section~\ref{sec:ufg}. These claims are that sets of cardinality one cannot be union-free and generic, and that only special cases of sets of cardinality two are union-free and generic. These two claims are proved in the following theorem. Furthermore, this theorem shows that the sets in $\Scal$ are connected in the sense that for every $S \in \Sscr$ with $\#S = m \ge 3$, there exists $S_m \in \Sscr$ such that $S_m \subsetneq S$ and $\#S_m = m-1$.

\begin{theorem}\label{th: S+card 2 and connect}
The family of sets $\Sscr$ given in Section \ref{sec:ufg} fulfills the following properties.
    \begin{enumerate}
        \item For every $p \in \Pcal$, $\{p\} \not\in \Sscr$.
        \item Let $\{p_1, p_2\} = S\in 2^\Pcal$. Then $S \not\in \Sscr$ if and only if the transitive reductions $tr(p_1)$ and $tr(p_2)$ differ only on one edge $(y_i,y_j)$ which is contained in either $tr(p_1)$ or $tr(p_2)$. This means that either $\#\{(tr(p_1)\setminus tr(p_2))\} = 1$ or $\#\{(tr(p_2) \setminus tr(p_1))\} = 1$ holds.
        \item $\mathscr{S}$ is connected in the sense that for every set $S\in\mathscr{S}$ of size $k\geq 3$ there exists a subset $S_m \subsetneq S$ of size $k-1$ that is in $\mathscr{S}$ too.
    \end{enumerate}
\end{theorem}
\begin{proof}
    Claim 1. follows directly from Condition (C1) of Definition~\ref{def:closure operator posets} as $\gamma(\{p\}) = \{p\}$ for every $p \in \Pcal$.

Now, we show the second claim. Let us first assume that $\{p_1, p_2\} = S \not\in \Sscr$. Using Part 1. we get that Condition~(C1) is not fulfilled. Hence, there exists no $p \in \Pcal$ such that $p \in \gamma(S) \setminus\{p_1, p_2\}$. Thus, the intersection must be either $p_1$ or $p_2$, (otherwise $p_1\cap p_2 \in \gamma(S)\setminus S$). W.l.o.g., let $p_1 = p_1 \cap p_2$. Then $p_2$ must be a superset of $p_1$ where there is no poset lying between $p_1$ and $p_2$. Therefore, $\#\{tr(p_2)\setminus tr(p_1)\} = 1$ is true.
    Conversely, assume that $S \in \Sscr$ and that $p_1$ is a superset of $p_2$. With this and Condition~(C1) we obtain that $\gamma(S) = \{p \in \Pcal\mid p_2 \subseteq p \subseteq p_1\}\subsetneq \{p_1, p_2\}$. Further assume that $\# \{tr(p_1)\setminus tr(p_2) \}= 1$ holds. However, with this, we get $\gamma(S) = S$ is true since no $p \in \Pcal$ can lie between $p_1$ and $p_2$. This is a contradiction which proves the claim.

Finally, we want to show the third part. Let $S\in \Sscr$ with $\#S \ge 3$. We show that there exists $S_m \in \Sscr$ with $S_m \subsetneq S$ and $\# S \setminus S_m = 1$. Let us distinguish the following two cases: 

\underline{\textit{Case 1:}} There exists $\tilde{p} \in S$ such that $D_{S}^{p, \cancel{\text{edge}}} \neq \emptyset$ for all $p \in S\setminus \tilde{p}$. Then we set $q_m = \cap_{p \in S\setminus \tilde{p}}\:  p$. Since for all $p \in S\setminus \{\tilde{p}\}$ we have $(M \times M) \setminus q_m \cap D_{S}^{p, \cancel{\text{edge}}} \neq \emptyset$, we can follow by Corollary~\ref{cor:ufg elements} that $q_m$ is an ufg element w.r.t. $S\setminus \{\tilde{p}\}$. Hence, $S_m = S\setminus \{\tilde{p}\} \in \Sscr$. 

\underline{\textit{Case 2:}} There exist $p_1, p_2 \in S$ such that $D_{S}^{p_i, \cancel{\text{edge}}} = \emptyset$ for $i \in \{1,2\}$. Before proceeding let us discuss the following claim:
%Note that $\emptyset\neq\tilde{S}\ni \{p_1,p_2\}$. In what follows we modify $\tilde{q}$ in such a manner that it is a ufg element of a proper subset of $S$ with only one less poset. Before that we have to discuss the following claim:

\textit{Claim$^*$:} Let $S \in \mathcal{S}$ with $\#S \ge 3$. Moreover, assume that there exist $\tilde{p_1}, \tilde{p_2} \in S$ such that $D_{S}^{\tilde{p}_i, \cancel{\text{edge}}} = \emptyset$ for $i \in \{1,2\}$. By Corollary~\ref{cor:ufg elements} there exists an ufg element $\tilde{q}$ which is given by Equation~(\ref{eq: ufg minimal element}). Analogously to this corollary, we set $\tilde{S} = \{p \in S\mid D_{S}^{p, \cancel{\text{edge}}} = \emptyset\} \supseteq \{\tilde{p}_1, \tilde{p}_2\}$. Then, we claim that there exists $p_0 \in \tilde{S}$ and $(a,b) \in \tilde{q} \cap D_{S}^{p_0, edge}$ such that there are no $y_1, \ldots, y_n \in M$ with $n \ge 3$ and $(a,y_1)(y_1, y_2), \ldots, (y_{n},b) \in tr(\tilde{q})$. In other words, $(a,b)$ cannot be given by a transitive chain and must be contained in the transitive reduction of $\tilde{q}$.

Assume in contradiction that Claim$^*$ is not true. Then for every $p_1 \in \tilde{S}$ and every element $(a,b)\in \tilde{q} \cap D_{S}^{p_1, edge}$ there must exists a sequence $y_1, \ldots, y_n \in M$ with $n \ge 3$ such that $(a,y_1)(y_1, y_2), \ldots, (y_{n},b) \in tr(\tilde{q})$.

Let $p_1 \in \tilde{S}$ and $(a,b)\in \tilde{q} \cap D_{S}^{p_1, edge}$. By assuming the contradiction, there exists $a=y^1_0, y^1_1, \ldots, y^1_{n+1} = b \in M$ such that $(y^1_0, y^1_1)(y^1_1, y^1_2) \ldots,(y^1_{n}, y^1_{n+1}) \in tr(\tilde{q})$. Since $\tilde{q}$ is in the style of Equation~\ref{eq: ufg minimal element}, the transitive reduction consists of elements that either lie in every poset or which lie in precisely one poset of $S$. Observe that every element $(y_{i-1}, y_i)$ for $i \in \{1, \ldots, n+1\}$ cannot lie in the intersection of all posets since then we have $(a,b)\in \cap_{p \in S} \: p$ which is a contradiction to $(a,b) \in \tilde{q} \cap D_{S}^{p_1, edge}$. Hence, there exists $i \in \{1, \ldots, n\}$ and $p_2 \in \tilde{S}$ such that $(y^1_{i-1},y^1_i)\in \tilde{q} \cap D_{S}^{p_2, edge}$. By assumption, we can again find a sequence of pairs in $tr(\tilde{q})$ such that $(y^1_{i-1},y^1_i)$ stems from transitivity. Note that $(a,b)$ cannot be an edge of the sequence to obtain $(y^1_{i-1},y^1_i)$ by transitivity, because then this leads to $\tilde{q}$ containing a cycle and not being antisymmetric. Again, this sequence contains at least one pair $(y^2_{j-1}, y^2_{j})\in \tilde{q} \cap D_{S}^{p_2, edge}$ and since we assume the contradiction of Claim$^*$ this pair can be obtained by a sequence representing the transitivity assumption. But now $(a,b)$ and $(y^1_{i-1}, y^1_i)$ cannot be used in the sequence to get $(y^2_{j-1}, y^2_{j})$. Since $M$ is finite this leads to a contradiction as the process needs to stop at some point. This proves Claim$^*$.

Let us go back to Case 2.  We set $\tilde{S} = \{p \in S\mid D_{S}^{p, \cancel{\text{edge}}} = \emptyset\}$. Since $S \in \Sscr$, there exists an ufg element $\tilde{q}$ w.r.t. $S$ which is given by Equation~(\ref{eq: ufg minimal element}). Now, we provide a procedure to give a modified version of $\tilde{q}$ (which we denote by $q_m$) which is then an ufg element w.r.t. a subset $S_m \subseteq S$ with $\# S\setminus S_m = 1$.
\begin{enumerate}
	\item Step 1: By Claim$^*$ there exists $p \in \tilde{S}$ and $(a,b) \in \tilde{q} \cap D_{S}^{p, edge}$ which cannot be divided by a transitive sequence. Hence  $\tilde{q} \setminus (a,b)$ is again a poset.
	\item Step 2: If $\{(a,b)\} = \tilde{q} \cap D_{S}^{p, edge}$, then we set $\tilde{q}_m = (\tilde{q}\setminus \{(a,b)\})$ and we are finished ($\tilde{q}_m$ is then the ufg element w.r.t. $S_m = S \setminus \{p\}$ and we can apply Lemma~\ref{lem:S_andere_def}). Else, we modify $\tilde{q}$ to $\tilde{q}^{next} = \tilde{q} \setminus (a,b)$. Note that $\tilde{q}^{next}$ is again an ufg element w.r.t. $S$ which is in the style of Equation~(\ref{eq: ufg minimal element}). This follows from the fact that there exists a further element $(\tilde{a}, \tilde{b}) \in \tilde{q} \cap D_{S}^{p, edge}$ with $(a,b) \neq (\tilde{a}, \tilde{b})$. Therefore, we can again apply Claim$^*$ on $\tilde{q}^{next}$ and go back to Step 1. This procedure will end after a finite number of steps since $M$ is finite. Hence, we get to a point where for one $p \in \tilde{S}$ and the modified $\tilde{q}^{next}$ we have $1 = \# (\tilde{q}^{next} \cap D_{S}^{p, edge})$ and can apply the first part of Step~2.
\end{enumerate}
%    The proof of the last part uses that the closure operator $\gamma$ stems from a formal context, which is a term from formal concept analysis. Since formal concept analysis is not part of this paper, we have outsourced the proof to \cite{Blocher_note}.
\end{proof}

Theorem~\ref{th: S+card 2 and connect} Part 1 gives us directly a lower bound for the cardinality of all sets $S \in \Sscr$. 
%Moreover, there exists a \textit{lower bound for all} $S \in\Sscr$, which is given by ${\#S\ge 2}$. This fact is already discussed in Example~\ref{example}.
For the upper bound, we use a complexity measure of $\Sscr$, the Vapnik-Chervonenkis dimension (VC dimension for short), see \cite{Vapnik15}. The VC dimension of a family of sets $\Cscr$ is the largest cardinality of a set $A$, such that $A$ can still be shattered into the power set of $A$ by $\Cscr$\footnote{To be more precise: The intersection between a set $A$ and a family of sets $\Cscr$ is defined by $A \cap \mathcal{C} = \{A \cap C \mid C \in \Cscr\}$. We say that a set $A$ can be shattered (by $\Cscr$) if $\# (A \cap \Cscr) = 2^{\# A}$ holds. The VC dimension of $\Cscr$ is now defined as $vc = \max \{\# A\mid (A \cap \Cscr) = 2^{ A}\}.$}.
With this, we obtain an upper bound for all $S \in \Sscr $ is given by ${\#S \le vc}$, with $vc$ the VC dimension of the closure system $\gamma(2^{\Pcal})$.
%The proof of the upper and lower bound can be found in Theorem \ref{th: upper, lower bounds S}. 
Note that in our case of posets, the VC dimension is small compared to the number of all posets.
\begin{theorem}\label{th: upper, lower bounds S}
For all $S \in \Sscr$, as defined in Section \ref{sec:ufg}, $\# S \ge 2$ and $\# S \le vc$ is true, where $vc$ is the VC dimension of the set $\gamma(2^{\Pcal})$.
\end{theorem}
\begin{proof}
Let $S \in \Sscr$. The proof for $\# S \ge 2$ follows immediately from Theorem \ref{th: S+card 2 and connect}.\\
To prove $\# S \le vc$ take an arbitrary subset $Q=\{p_1,\ldots, p_k\}\in \Sscr$ of size $k> vc$. Then this subset is not shatterable because $vc$ is the largest cardinality of a shatterabel set. Thus there exists a subset $R\subseteq Q$ that cannot be obtained as an intersection of $Q$ and some $\gamma(A)$ with $A \subseteq \Pcal$. In particular, this holds for $R = A$. Thus, $R \neq \gamma(R) \cap Q$ and with the extensitivity of $\gamma$ we get $R \subsetneq \gamma(R)\cap Q$. This means that there exists an order $p_i$ in $\gamma(R) \cap Q \backslash R$ for which the formal implication %with $\gamma(R) =\gamma(R\backslash \{p_i\})$ which means that the formal implication
$R\rightarrow \{p_i\}$ holds. Thus, (because of the Armstrong rules, cf., \cite[p. 581]{armstrong74}) the order $p_i$ is redundant in the sense of $Q\backslash\{p_i\} \rightarrow Q$ and thus $Q$ is not minimal with respect to $\gamma$. Therefore, $Q$ is not in $\Sscr$ which completes the proof.
\end{proof}
\begin{remark}\label{rem: upper,observed, vc bound}
In concrete applications, one usually does not observe all possible posets. Therefore, many summands in the definition of the empirical ufg depth (see Definition~\ref{def: empirical ufg depth}) are zero. Thus, one can restrict the analysis to $\Sscr_{obs}:= \{ S \in \Sscr \mid \mbox{ all } p \in S \mbox{ are observed} \}.$ Then a similar argumentation shows that $\#S\leq vc_{obs}$ where $vc_{obs}$ is the VC dimension of $\gamma(2^{\mathcal{P}_{obs}}) \cap \mathcal{P}_{obs}$ and $\mathcal{P}_{obs}$ is the set of all observed posets.
\end{remark}
Beyond the bound for the size of a set $S \in \Sscr$ that is given by the VC dimension and that is generally valid, we can additionally give another bound for the special case of poset data that is sharper than the VC bound in certain cases.
\begin{theorem}\label{tighterbound}
    Let $m:= \#M \geq 3$. Then, for every $S \in \Sscr$ we have that $\# S \leq m(m-1)/2$.
\end{theorem}
\begin{proof}
    Let $M=\{x_1,x_2,\ldots,x_m\}$. Let $S=\{p_1,\ldots,p_{\ell}\} \in \Sscr$ and let $q$ be an ufg element w.r.t. $S$. Then, by Corollary~\ref{cor:ufg elements}, for all $p \in S$ we have that $q \cap D_{S}^{p, \text{edge}} \neq \emptyset$ or $(M \times M) \setminus q \cap D_{S}^{p, \cancel{\text{edge}}} \neq \emptyset$. This means that for every poset in $S$ there is an edge $(x_i, x_j) \in p$ with $i, j \in \{1, \ldots, m\}$ and $i \neq j$ such that $p$ is needed to either
 \begin{itemize}
 	\item[(i)] contribute in $q$ by providing an edge $(x_i, x_j)\in q \cap p$ that no other poset in $S$ has, or
 	\item[(ii)] to contribute in $q$ by not having an edge $(x_i, x_j)\notin q$ and $(x_i, x_i)\notin p$ that every other poset has.
 \end{itemize}
 This means that $p$ has a unique existence-characteristic w.r.t. edge $(x_i, x_j)$ and $S$, while all other posets in $S$ agree to have the opposite existence-characteristic w.r.t. $(x_i, x_j)$ and $S$. That is, if $(x_i, x_j) \in p$, for all $\tilde{p} \in S \setminus\{p\}$ we have $(x_i, x_j) \not\in \tilde{p}$ and vice versa for $(x_i, x_j) \not\in p$. \\
Since by Corollary~\ref{cor:ufg elements} each poset must somehow contribute uniquely to $q$, we get that each edge can be used by only one single poset in $S$. This follows from the fact that $\#S \ge 3$ and all other posets $\tilde{p} \in S \setminus\{p\}$ need to agree on the opposite existence-characteristic w.r.t. $(x_i, x_j)$. With this we obtain that $\#S \le m(m-1)$ (since the reflexive part holds by default for every poset). \\
We continue with the above poset $p\in S$ and the corresponding edge $(x_i, x_j)$ that uniquely contributes to $q$ by $p$. We show that the inverse edge $(x_j, x_i)$ cannot be a contributing element for any $\tilde{p} \in S \setminus \{p\}$. To show this, we split the proof into two cases:\\
 \underline{Case 1:} $(x_i, x_j) \in q \cap D_{S}^{p, \text{edge}}.$ This means that all posets $\tilde{p}\in S \setminus \{p\}$ agree on $(x_i, x_j) \not\in \tilde{p}$. We get that any poset $\tilde{p}\in S \setminus \{p\}$ cannot use $(x_j, x_i) \in \tilde{p}$ to uniquely contribute to $q$, since this edge never occurs in $q$ ($q$ is a poset and therefore antisymmetric). Moreover, since $(x_j, x_i) \not\in p$ is true, $\tilde{p}$ cannot contribute uniquely to $q$ by $(x_j, x_i) \not\in \tilde{p}$. Since $\#S \ge 3$, there cannot be another poset $\tilde{p}$ that uses the inverted edge as a unique contribution to $q$. \\
 \underline{Case 2:} $(x_i, x_j) \in (M \times M) \setminus q \cap D_{S}^{p, \cancel{\text{edge}}}$. With this we get that for all posets $\tilde{p}\in S \setminus \{p\}$ we have $(x_i, x_j) \in \tilde{p}$. Due to antisymmetry we immediately get that $(x_j, x_i) \not\in \tilde{p}$ for all $\tilde{p} \in S \setminus \{p\}$. With $\#S \ge 3$ we have that $(x_j, x_i)$ cannot be used to uniquely contribute to $q$ by a poset $\tilde{p}$. \\
 With these two cases we have that only half of all possible edges can be used and therefore we have shown that $\# S \leq m(m-1)/2$.
\end{proof}

\begin{remark}
    The bound of Theorem~\ref{tighterbound} is tight in the sense that there exists a set $S\in\Sscr$ with cardinality $m(m-1)/2$, namely the set $$S:=\{\{(x_i,x_j)\}\mid i,j \in \{1,\ldots, m\},  i<j\}.$$
\end{remark}

We conclude with a technical observation that we need to analyze the properties of the (empirical) ufg depth. The next lemma states that the set $\Sscr$ can be rewritten.
\begin{lemma}\label{th: restructuring Sscr}
	For $p \in \Pcal$ we get
        \begin{align}\label{eq:Sscr als Schnitt_1}
	&\{S\in \Sscr\mid p \in \gamma(S)\} \\
	&= \bigcap\limits_{(y_i, y_j) \in p}\{S\in \Sscr\mid \exists p \in S\colon (y_i, y_j) \in p\}\: \cap \label{eq:Sscr als Schnitt_2}\\
	&\quad \bigcap\limits_{(y_i, y_j) \not\in p}\{S\in \Sscr\mid \exists p \in S \colon (y_i, y_j) \not\in p\} \label{eq:Sscr als Schnitt_3}.
\end{align}
\end{lemma}
\begin{proof}
Let $p \in \Pcal$. The proof is divided into two parts. \\
Part 1: We prove $\subseteq$. Let $S$ be an element of (\ref{eq:Sscr als Schnitt_1}). Since $p \in \gamma(S)$, we have $p \subseteq \cup_{\tilde{p} \in S} \:\tilde{p}.$ So for every $(y_i,y_j) \in p$ there is a $\tilde{p} \in S$ such that $(y_i, y_j) \in \tilde{p}$. Therefore, $S$ is an element of the intersection of (\ref{eq:Sscr als Schnitt_2}). Also from $p \in \gamma(S)$ we get $\cap_{\tilde{p} \in S}\:\tilde{p} \subseteq p$ and thus we know that for every $(y_i,y_j) \not\in p$ there exists a $\tilde{p} \in S$ such that $(y_i, y_j) \not\in \tilde{p}$. Thus, $S$ is an element of the intersection given by (\ref{eq:Sscr als Schnitt_3}). This proves Part 1.\\
Part 2: We prove $\supseteq$. Let $S\in \Sscr$ be an element of the right-hand side of the equation. We show that $p \in \gamma(S)$. Let $S$ be in the intersection given by (\ref{eq:Sscr als Schnitt_2}). Then we know that for every $(y_1,y_2) \in p$ there exists an $\tilde{p}\in S$ such that $(y_1, y_2) \in \tilde{p}$. Thus $p \subseteq \cup_{\tilde{p} \in S}\: \tilde{p}$. The second part of the intersection given by (\ref{eq:Sscr als Schnitt_3}) analogously yields that $\cap_{\tilde{p}\in S} \: \tilde{p} \subseteq p$. Hence $p \in \gamma(S)$ and the second part is proven. The claim follows from Part 1 and Part 2.
\end{proof}

\subsection{Properties of the (Empirical) UFG Depth}\label{sec: prop_ufg_depth}
In this subsection, we introduce properties of the (empirical) probability measure $D_n$ and $D$.
The following statements focus on $D_n$. Those properties that use only the empirical measure and not the concrete sample values can be transferred to $D$.

The first observation is that the ufg depth considers the orders as a whole, not just pairwise comparisons. More precisely, the ufg depth cannot be represented as a function of the sum-statistics $$\left(w_{(a,b)} := \# \{i \in\{1,\ldots,n \} \mid (a,b) \in p_i\}\right)_{(a,b) \in M \times M}$$ of the pairwise comparisons, see Theorem~\ref{thm_pairwise}. Note that many classical approaches rely only on the sum-statistics. For example, within the Bradly-Terry-Luce model (cf.,~\cite[p. 325]{btl}) or the Mallows $\Phi$ model (cf.,~\cite[p. 360]{fligner}), the likelihood function that is maximized depends only on the data through the sum-statistics. %,  see \cite{btl}. Another example would be the case where one defines the center of a set of posets as that order that minimizes the sum of the Kendalls tau distance (cf., \cite{fagin06}) to all observed orders.}%  a metric based data depth notion where one looks at the kendalls distance of the uses with a fixed centerthat reonly the sum-statistics the Likelihood One can use distance measures, and then the poset with the minimum mean distance defines the central observation. Note that many distance measures, such as Kendall's tau (see \cite{fagin06}), are based on pairwise comparisons. EVtl.: \cite{BTL} [morgen besprechen]}

\begin{theorem}\label{thm_pairwise}
    $D_n$ cannot be represented as a function of the sum-statistics $w_{(a,b)}$.
\end{theorem}
\begin{proof}
    We simply give two data sets $\mathcal{D}=(p_1,p_2,p_3)$ and $\tilde{\mathcal{D}}=(\tilde{p}_1,\tilde{p}_2,\tilde{p}_3)$ on the basic set $M=\{y_1,y_2,y_3\}$ with the same sum-statistics but different associated depth functions: Let $p_1,p_2$ and $p_3$ be given as the transitive reflexive closures of $\{(y_1,y_2)\}$;  $\{(y_1,y_2),(y_1,y_3)\}$ and $\{(y_2,y_3), (y_1,y_3)\}$, respectively. Let  $\tilde{p}_1$, $\tilde{p}_2$ and $\tilde{p}_3$ be the transitive reflexive closure of $\{(y_1,y_2)\}$; $\{(y_1,y_3)\}$ and $\{(y_1,y_2),(y_2,y_3)\}$, respectively. Then both data sets have the same sum-statistics $w_{(y_1,y_2)}=w_{(y_1,y_3)}=2$; $w_{(y_1,y_3)}=1$ and $w_{(y_i,y_j)}=0$ for all other $y_i\neq y_j$. But the ufg depth of $p_1=\tilde{p}_1$ is $1/2$ w.r.t. the first data set but $7/10$ w.r.t the second data set. The corresponding code can be found at the link mentioned in Footnote~\ref{footn: github}.
\end{proof}

Remarkably, this concretization by Theorem~\ref{thm_pairwise} formalizes precisely the analogy to the ontic notion of non-standard data mentioned at the beginning: Computing the depth of a partial order cannot be broken down via simple sum-statistics, but requires the partial order as a holistic entity. This is due to the fact that the involved set operations within the closure operator $\gamma$ rely on the partial orders as a whole.

In Section \ref{sec:ufg}, we defined the ufg depth in terms of two cases. If there exists at least one element $S \in \Sscr$ such that every $p \in S$ has a positive empirical measure, then $D_n \not\equiv 0$. In Corollary~\ref{cor: Dn always zero} we specify this non-triviality property. We claim that $D_n \equiv 0$ occurs only when either the entire (empirical) probability mass lies on one poset or when the (empirical) probability mass is on two posets where the transitive reduction differs only in one pair, see Theorem~\ref{th: S+card 2 and connect}.
\begin{corollary}\label{cor: Dn always zero}
   $D(p) = 0$ for every $p \in \Pcal$ if and only if the measure $\nu$ has either the entire positive probability mass on a single poset $p$ or on exactly two posets $p_1$ and $p_2$ where the transitive reduction differs only in a pair $(y_1,y_2)$,i.e., either $\#\{tr(p_1) \setminus tr(p_2)\} = 1$ or $\#\{ tr(p_2)\setminus tr(p_1)\} = 1$.
\end{corollary}
\begin{proof}
    Note that $D(p) = 0$ for every $p \in \Pcal$ is true if for all $S \in \Sscr$, $\prod_{\tilde{p} \in S}\nu(\tilde{p}) = 0$. Theorem \ref{th: S+card 2 and connect} Part 1. and 2. provide the cases when this holds which proves immediately the claim.

    The converse follows analogously from Theorem \ref{th: S+card 2 and connect}.
\end{proof}

The next observation relates to how the sampled posets affect the ufg depth value. Therefore, let us recall Example~\ref{example} and Example~\ref{example_2}. From the structure of the sample, we can immediately see that $p_\Delta$ has a nonzero depth and that $p_{total}$ must have a depth of zero. Now, let us take a closure look at how the structure of the sample effects $D_n$. Therefore, let $\underline{p}= (p_1, \ldots, p_n)$ be a sample from $\Pcal$. Let $(y_1,y_2) \in M\times M$ such that for all $i \in \{1, \ldots, n\}, \: (y_1,y_2) \not\in p_i$. Then for every $p\in \Pcal$ with $(y_1,y_2) \in p$, we get $D_n(p) = 0$. This means that if a pair does not occur in any poset of the sample, then every poset that contains this pair needs to have zero empirical depth.
%such that Then the empirical depth value of a poset $p$ with %$(y_i, y_j) \in p$ but for all $i \in \{1, \ldots, n\}$, $(y_i, y_j) \not\in p_i$ is $D_n(p) = 0$.$(y_i, y_j) \in p$ but for all $i \in \{1, \ldots, n\}$, $(y_i, y_j) \not\in p_i$ is $D_n(p) = 0$. 
Reverse, when looking at non-pairs, a similar statement is true. Let $p \in \Pcal$ with $(y_1, y_2) \not\in p$ but for all $i \in \{1, \ldots, n\}$, $(y_1, y_2) \in p_i$ holds. Then, $D_n(p) = 0$.
This follows from Corollary \ref{cor: Dn and the sample}.
This discussion is based on whether a pair has been observed or not. Now, we are interested in how duplicates influence the value of the empirical ufg depth ${D_n}$. This is immediately apparent by using the empirical measure $\nu_n$. Thus, each element in $\Sscr$ is weighted by the number of duplicates in the sample $\{p_1, \ldots, p_n\}$.

Conversely to the question of how the sample effects the values of $D_n$, in some cases, structure in the sample can be inferred by the ufg depth values. In Example~\ref{example} and Example~\ref{example_2}, knowing only the values of the depth function gives us some insight into the observed posets. For example, we know that there must be at least one pair $(y_i, y_j)$ that is an element of $p_{total},$ but which is not given by any observed poset. Moreover, the fact that $p_\Delta$ has nonzero depth implies that there exists no pair $(y_i, y_j)$ that every observed poset has. More precisely, the depth value of the trivial poset, which consists only of the reflexive part, as well as the values of the total orders, can provide further information about the sample. Therefore, let $p_{\Delta}$ be the trivial poset, and $p_{\text{total}}$ be a total order. This implication of the outliers on the sample property is discussed by Corollary~\ref{cor: Dn and the sample}. 
%This corollary gives us that if $D_n(p_{\Delta}) = 0$, then there exists at least one pair $(y_1, y_2)$ which is in every poset of the sample. The knowledge of $p_{total}$ leads to an statement about the non-edges. So, if $D_n(p_{\text{total}}) = 0$ is true, then there exists at least one pair $(y_1, y_2) \in p_{\text{total}}$ which is in no poset of the sample.

\begin{corollary}\label{cor: Dn and the sample}
	Let $(p_1, \ldots, p_n)$ be a sample of $\Pcal$. Let $\nu_n$ be the empirical probability measure induced by $(p_1, \ldots, p_n)$. Furthermore, let $\nu_n$ be such a probability measure that $D_n \not\equiv 0$. Then for $D_n$, the following statements are true.
	\begin{enumerate}
		\item Assume that for all $p_i \in \{p_1, \ldots, p_n\}$, $(y_1, y_2) \in p_i$ is true. Then for every poset $p \in \Pcal$ with $(y_1, y_2) \not\in p$, $D_n(p) = 0$ follows.
		\item Assume that for all $p_i \in \{p_1, \ldots, p_n\}$, $(y_1, y_2) \not\in p_i$ holds. Then for every poset $p \in \Pcal$ with $(y_1, y_2) \in p$, $D_n(p) = 0$ is true.
		\item Let $p_{\Delta}$ be the poset consisting only of the reflexive part. $D_n(p_{\Delta}) = 0$ if and only if there exists a pair $(y_1, y_2)$ such that for all $p_i \in \{p_1, \ldots, p_n\}$, $(y_1, y_2) \in p_i.$
		\item Let $p_{total} \in \Pcal$ be a total order. $D_n(p_{total}) = 0$ if and only if there exists a pair $(y_1, y_2)\not\in p_{total}$ such that for all $p_i \in \{p_1, \ldots, p_n\}, (y_1, y_2) \in p_i$ is true.
	\end{enumerate}
\end{corollary}
\begin{proof}
    First, note that for $S \in \Sscr$, where there exists an $\tilde{p} \in S$ such that $\nu_n(\tilde{p}) = 0$, $S$ contributes nothing to $D_n$. So one can replace $\Sscr$ in the definition of $D_n$ by $\Sscr_{obs}$, see Remark~\ref{rem: upper,observed, vc bound}. The reduced set $\Sscr_{obs}$ is used to show the claims.

    The proof of Claim 1., 2., 3. and 4. are analogous. Hence, here we provide only the proof of Claim~1. Let $(y_1, y_2) \in M \times M$ such that for all $i \in \{1, \ldots, n\}$ $(y_1,y_2) \in p_i$ and let $p \in \Pcal$ such that $(y_1, y_2) \not\in p$. Let $S \in \Sscr_{obs}$ and take a closer look at Equation-part~(3) of Lemma \ref{th: restructuring Sscr}. Since $(y_1,y_2) \not\in p$, $S$ cannot be an element of the intersection of (3). Thus, $\{S \in \Sscr_{obs} \mid p\in\gamma(S)\}$ is empty and with the comment above we get that $D_n(p) = 0$.
    %Consider (3) of Theorem \ref{th: restructuring Sscr}. Thus, every $S$ which is an element of the intersectionwithin in the intersection given by (3) must have at least one partial order $p$ where $(y_1, y_2) \not\in p$ is true. However, these $S$'s cannot lie in $\Tilde{\Sscr}$ and hence the depth of $D_n(p) = 0$.
\end{proof}

The last properties have summarized how the structure of a sample is reflected in the ufg depth and vice versa. Finally, we take a look at the consistency of the empirical ufg depth $D_n$. This means that $D_n$ converges uniformly to $D$ almost surely under the assumption of observing i.i.d. samples, see Theorem~\ref{th: convergence}. 

\begin{theorem}\label{th: convergence}
    Let for all $n \in \mathbb{N}$ the sample $(p_1, \ldots, p_n)$ be i.i.d. according to distribution $\nu \in \mathcal{M}$. Then the corresponding empirical depth function $D_n$ almost surely uniformly to $D(\cdot, \nu)$ for $n$ going to infinity.
\end{theorem}
\begin{proof}
    Due to the i.i.d. assumption and the law of large numbers, we know that for every $p \in \Pcal, \:\left\lVert \nu_n(p) - \nu(p)\right\rVert \overset{n \to \infty}{\to} 0$ almost surely (a.s). Since $\#\Pcal$ is finite, we get that $\nu_n$ also converges a.s. uniformly to $\nu$. Finally, we use that $D_n$ and $D$ are both the same finite composition of $\nu_n$ and $\nu$, respectively, and we obtain $
    \sup_{p \in \Pcal} \left\lVert D_n(p) - D(p)\right\rVert \overset{n \to \infty}{\to} 0 \text{ almost surely.}$
\end{proof}

\section{Implementation}\label{subsec:implementation}
In this section, we discuss the difficulties and corresponding solution approaches in computing $D_n$.
Therefore, let $\underline{p} = (p_1, \ldots, p_n)$ be a sample of posets. 
The naive approach is to just check all subsets of $\{p_1, \ldots, p_n\}$ whether they are in $\Sscr_{obs}:= \{ S \in \Sscr \mid \mbox{ all } p \in S \mbox{ are observed} \}$ is very time-consuming, especially since the subsets that are elements of $\Sscr_{obs}$ can be very sparse in $2^{\{p_1, \ldots, p_n\}}$. Moreover, it is difficult to test whether a subset is an element of $\Sscr_{obs}$ or if it is not an element by taking Conditions~(C1) and~(C2) as a basis. Finally, even if we are able to compute the entire family $\Sscr_{obs}$, computing all possible poset $p$ on a set $M$ to obtain some insights on $D_n$ is for even small $M$ with $\# M \ge 8$ very computation intensive and for larger $\# M$ it is currently not possible.\footnote{Even computing the number of possible posets is not feasible for $\#M > 18$, see \url{https://oeis.org/A001035} (Accessed 10.11.2023).} We address all these issues in the next paragraphs.

Let us start with computing $\Sscr_{obs}.$ First, we can use the lower and upper bound on the cardinality of $S \in \Sscr_{obs}$, see Theorem~\ref{th: upper, lower bounds S} and Remark~\ref{rem: upper,observed, vc bound}. Here we use the binary linear programming formulation described in \cite[p.33f]{epub40416} to compute the VC dimension. Further, we use the connectedness of the elements $S \in \Sscr$, see Theorem~\ref{th: S+card 2 and connect} Part 3. With this, we do not have to go through every subset that lies between the lower and upper bounds, but can stop the search earlier.

Still, we need to check whether a subset $P \subseteq \{p_{1} \ldots, p_{n}\}$ is an element of $\Sscr_{obs}$. Therefore, we use Corollary~\ref{cor:ufg elements} together with Definition~\ref{def: ufg element} and~\ref{def:disting}. Hence, it is sufficient to test if an ufg element w.r.t. to $P$ that is in the style of Equation~(\ref{eq: ufg minimal element}) exists. Based on this, we start with computing $D_{P}^{p, \text{edge}}$, $D_{P}^{p, \cancel{\text{edge}}}$ and $\tilde{S} = \{p \in P \mid D_{P}^{p, \cancel{\text{edge}}} = \emptyset\}$. With this, we can compute all $\tilde{q}$ that is in the style of Equation~(\ref{eq: ufg minimal element}). If one $th(\tilde{q}) \subseteq \cup_{p \in P} \: p$ is true, then $P \in \Sscr_{obs}$ since $\tilde{q}$ is an ufg element w.r.t. $P$ by Lemma~\ref{lem:S_andere_def}. Note that the reverse that $\cap_{p \in P}\: p \subseteq th(\tilde{q})$ follows directly from the definition given by $\tilde{q}$ via Equation~(\ref{eq: ufg minimal element}).

Now we achieved to compute the whole family $\Sscr_{obs}$. The next issue we approach is that computing all possible posets $p$ on $M$ can be difficult. For small $\#M$, say up to $6$ or $7$, this is computationally tractable. In this situation, we exploit the fact that every partial order can be represented as the intersection of its linear extensions. Furthermore, we use that the set of all partial orders, together with the relation $M \times M =\{(x,y)\mid x,y \in M\}$, which consists of every element of $M \times M$, defines a closure system on $M\times M$. This allows us to use efficient algorithms from formal concept analysis.\footnote{
For a deeper understanding of this computational approach, knowledge of formal concept analysis is required, see~\cite{ganter12}. This implementation is based on a so-called formal context $(O,A,I)$, where each object $o \in O$ is a linear order $L$ (on $M$) and each attribute $a \in A$ is a pair $(x,y)\in M\times M$ and $(L, (x,y)) \in I \iff (x,y) \in L$. Then, since every partial order is an intersection of a set of linear orders (more concretely, the set of all its linear extensions), the intents of this context are exactly all partial orders on the set $M$ (plus the relation $M\times M$). Therefore, we can compute the set of all partial orders by computing the intents of this formal context instead, e.g. with the next closure algorithm, see \cite{ganter2010two}. }

Having explained how we compute all possible partial orders when it is tractable, we now discuss the case where computing all possible posets is no longer feasible. In this case, we use a binary linear program that gives us maximum and minimum depth values, as well as representative posets for each of these values. The binary linear program is defined as follows
\begin{align}\label{milp}
    \sum_{S \in \Sscr_{obs}} \frac{w_S}{c_n} \cdot x_S &+ \sum_{(a,b) \in M\times M} 0 \cdot x_{(a,b)} \longrightarrow \max \\%\nonumber 
     \text{subject to }&  A_{poset} \: x\ge b_{poset} \nonumber \\ 
    & A_{intersect} \: x \ge b_{intersect} \nonumber \\ 
    & A_{union} \: x \ge b_{union} \nonumber \\
    & x_i \in \{0,1\} \quad \text{for}\: i \in  \Sscr_{obs} \cup (M \times M) \nonumber\nonumber
\end{align}
where $\frac{w_S}{c_n}$ for $S \in \Sscr_{obs}$ are the weights given by duplicates and the cardinality of $S$ together with the normalization constant, see Definition~\ref{def: empirical ufg depth}. %and $x_i \in \{0, 1\}$ for $i \in \Sscr_{obs} \cup (M \times M)$. 
The first $\#\Sscr_{obs}$ elements of $x$ describe the union-free generic set, and the next ${\#M \cdot \#M}$ elements guarantee that we only discuss posets. First, we need to ensure that only relations that define a poset are discussed. This is represented in Program~(\ref{milp}) by the constraint $A_{poset} \: x\ge b_{poset}$. Therefore matrix $A_{poset}$ together with $b_{poset}$ represents only the reflexive, transitive, and antisymmetric constraints on $x_{(a,b)}$ with $(a,b) \in M \times M$. So in this part, the matrix entries of the first $\#\Sscr_{obs}$ columns are all zero. Second, we have to include in the binary linear program that a set $S \in \Sscr_{obs}$ can only be included in the maximum if the considered poset lies in the intersection of all posets of $S$, see $A_{intersect} \: x \ge b_{intersect}$. Each row of this constraint corresponds to one ufg set $S \in \Sscr$ and imposes a structure on $x_{(a,b)}$ with $(a,b) \in M \times M$ when this set $S$ is included in the maximization of the objective function. Similarly, we ensure with $\quad A_{union} \: x \ge b_{union}$ that a set $S \in \Sscr_{obs}$ is only counted if the poset considered is a subset of the union of all posets in $S$. Now, computing the maximum leads to the highest depth value and a poset that has that highest depth value. However, if we are interested in the smallest depth value, this binary linear program does not work, because we have not yet enforced that a set $S \in \Sscr_{obs}$ which has the currently analyzed poset in its closure must be represented in the calculation of the objective function. Therefore, we need to include further constraints so that when a poset is used to minimize the objective, every set $S \in \Sscr_{obs}$ that has that poset in its closure is counted in the objective sum. The code with detailed comments can be found on GitHub, see Footnote~\ref{footn: github}.

%$A_{union} \: x \le b_{union}$ has $\#\Sscr$ rows and each single row ensures if one single $S \in \Sscr$ counts to the maximal sum that the corresponding poset is a subset of the union of the pairs of $\Sscr$. $A_{intersect} \: x \le b_{intersect}$ aims the reverse. It ensures that the the poset is a superset of the intersection. Now, we have implemented all conditions of the closure operator, see Definition~\ref{def:closure operator posets}. Finally, we have to guarantee that only posets are included. Thus, we implement the reflexivity, transitivity and antisymmetry assumption to $\quad A_{poset} \: x\le b_{poset}$. With Program~\ref{milp} we compute the maximal depth value $d_{max}$ together with a poset which has this maximal depth value. If we include further the constraint that the already observed posets are not allowed to be used for the maximal poset computation, we can compute iterative all $k$ deepest ufg depth values and always one representative poset which has this depth value. For further information on the implementation, we refer to the implementation, see Footnote~1. \textcolor{red}{@Hannah Nebenbedingungen raus und nur erklären, dafür mehr und warten bis Georg implementiert hat um Rest zu erklären. In der Zielfunktione fehlt noch gewichtung. Wir nennen das ganze binary linear program}

All in all, we improved the computation compared to the naive approach by using the knowledge provided in Section~\ref{sec:properties}. 
By the two bounds on the cardinality of the sets in $\Sscr_{obs}$ we get an idea of the worst and best case of the computation time. By further using Lemma~\ref{lem:S_andere_def} with Corollary~\ref{cor:ufg elements} 
%instead of Conditions (C1) and (C2) 
together the connectedness property, see Theorem~\ref{th: S+card 2 and connect} Part 3, we could decrease the computation time. Although we currently cannot calculate the exact amount of this reduction in general as this depends on the complexity of the data set used. Note that the upper bound using the VC dimension is not fixed, but depends on the structure of the data set. However, the bound given by Theorem~\ref{tighterbound} holds in general, but can be quite loose in some data situations. Finally, by stating a binary linear program, we achieved that even if not all posets are computable, we get the highest and lowest ufg depth values with representative posets. \ref{appendix: computation time} summarizes the computation time and complexity of the two examples in the next section.

\section{Application on Classifier Comparison} \label{sec:application}
%Moreover, we analyzed the connection between the observed posets and the ufg depth values. 
In the above sections, we have discussed theoretically how the union-free generic depth captures the structure of the partial orders and how it can be computed. 
Now, we take the next step and apply our ufg depth to poset-valued data, where each poset arises from the comparison of machine learning algorithms based on multiple performance measures on data sets. Concretely, let us consider $k$ algorithms which are evaluated on the basis of $\ell$ performance measures on $n$ different data sets. Thus, for each data set and each algorithm, we have $\ell$ performance measures. This allows us to compare all algorithms based on a single data set. We say that algorithm $i$ is better than/outperforms algorithm $j$ if and only if there exists at least one performance measure such that algorithm $i$ is strictly better than algorithm $j$ and for all other performance measures algorithm $i$ does not perform worse than algorithm $j$. If two performance measures contradict each other in the sense that for one measure algorithm $i$ is strictly preferred and for another measure algorithm $j$ is strictly preferred, we say that the two algorithms are incomparable. This gives us a poset for each data set, which describes the performance of the algorithms based on that data set. Thus, we observe in total $n$ posets.\footnote{It may occur that two algorithms are equal on all performance measures. In this situation, the two algorithms are indifferent based on the performance measures used. Thus, the described procedure does not lead to a partial order, but only to a preorder. Note that in this situation it is not appropriate to say that there is no dominance structure between these two algorithms, and therefore to include this as one of the incomparability parts in the poset. This confuses the distinction between incomparability and indifference. Here we assume that there are only incomparabilities and no indifference. In the following, we restrict our analysis to algorithms that are substantially different and therefore do not produce the same performance measures.} In what follows, we focus on benchmarking classification algorithms.

We provide two examples for comparing machine learning algorithms based on ufg depth. Both examples use openly available repositories containing data sets with binary classification problems. For each data set in the repository, multidimensional performance measures exist, and in this paper, we use these measures to obtain the corresponding posets. 
%Thus, considering 80 data sets, we obtain 80 posets describing for each data set the performance structure of the classifier algorithms of interest. 
The aim of the first example is to highlight the difference between our approach and other approaches analyzing poset-valued data. The second example demonstrates how the ufg method can be used to gain insight into the performance of different algorithms.

\subsection{UCI Repository: Comparison to Related Methods}\label{sec: uci}
In this section, we illustrate the difference between our approach and other existing approaches to analyze poset-valued data. We focus on an extension of the Bradley-Terry model to ties, see~\cite{btl, davidson70}, and an approach based on generalized stochastic dominance, see~\cite{jansen23}. Beyond these two approaches, there are others, such as the Plackett-Luce model, see~\cite{plackett75, baker21}, which are not discussed here. We use the data sets provided by the openly available UCI repository, see~\cite{dua17}, to obtain the poset-valued data set.
%discussed in~\cite{jansen23} to demonstrate the different aspects of these approaches.

\subsubsection{Data Set}
We take 16 data sets from the UCI machine learning repository, see~\cite{dua17}. 
%These data sets compare classifiers based on their performance. 
They all focus on classifier comparison and vary greatly in size, dimensionality, and class imbalance. The following poset-valued data describing the performance of classifiers is based on the performance evaluation and analysis in~\cite{jansen23}.

We are interested in the following supervised learning methods: \textit{Boosted Decision Stumps} (BS), \textit{Decision Tree} (CART), \textit{Elastic net penalized logistic regression} (EN), \textit{Gradient Boosting} (GBM), \textit{Generalised Linear Model} (GLM), \textit{L1 penalized logistic regression} (LASSO), \textit{Random Forest} (RF) and \textit{L2 penalized logistic regression} (RIDGE). Their performance is compared using \textit{predictive accuracy, area under the curve} and \textit{Brier score}. These performance measures were calculated by~\cite{jansen23}. We refer to \cite{jansen23} (Section 6.1 and Appendix A.2) for more details on the data set selection, implementation, and evaluation of the performance measures.

\begin{figure}
    \centering
    \includegraphics[scale = 0.6]{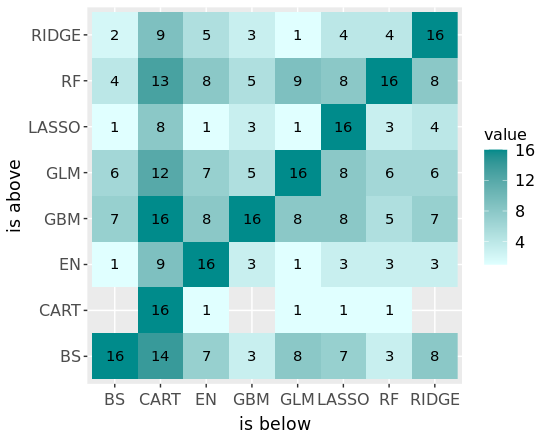}
    \caption{UCI: Heatmap representing the sum-statistics, see Section~\ref{sec:properties}.}
    \label{fig:heatmap_uci}
\end{figure}

%For each data set, we obtain one poset describing the performance order of the eight classifiers. 
By using the procedure given at the beginning of Section~\ref{sec:application}, 
%and the three performance measures evaluated for each classifier and each data set, 
we obtain one poset describing the performance structure of the eight classifiers for each data set.
%We say that, given a data set, classifier $i$ outperforms classifier $j$ if and only if there exists at least one performance measure such that classifier $i$ is strictly preferred over classifier $j$, and for all other performance measures, classifier $i$ does not perform worse than classifier $j$. Otherwise, we say that these two classifiers are incomparable. Since we consider 16 data sets here, we obtain 16 posets describing the performance of the eight classifiers. 
Thus, we observe 16 posets. These posets have no duplicates. Figure~\ref{fig:heatmap_uci} shows the heatmap of the pairwise comparison of the classifiers based on all three performance measures. We can observe that all posets agree that CART performs worse (in all dimensions) than GBM. We can also see that CART never dominates BS and RIDGE. However, only 14 (respectively nine for the comparison with RIDGE) posets state that BS (respectively RIDGE) is better than CART. For all other posets, CART and BS (or RIDGE) are incomparable in the sense that two performance measures give an opposite evaluation of their performance. Note that the diagonal of Figure~\ref{fig:heatmap_uci} needs to be 16 as all posets are reflexive.

Looking only at the pairwise comparisons, there exists no classifier that clearly dominates all other classifiers. At first glance, it seems that CART has the lowest performance.

\subsubsection{Discussion on Related Methods}

\begin{figure}
    \centering
    \includegraphics[width=\linewidth]{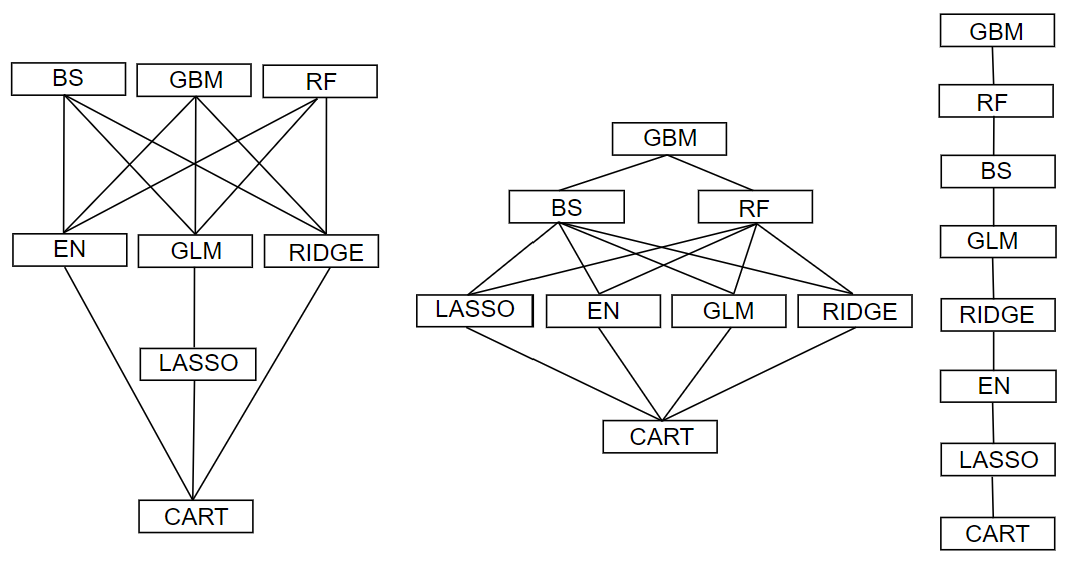}
    \caption{UCI: Poset having maximal depth value based on all possible posets (left), the poset given by the generalized stochastic dominance approach (middle), see~\cite{jansen23}(Figure 5 upper graph), and the ranking given by the extended Bradley-Terry model (right). E.g. for all three posets we have that CART is dominated by all other classifiers.}
    \label{fig:uci_max_bt}
\end{figure}

The ufg depth approach presented in this paper provides a depth measure for all possible partial orders given by the items \textit{BS, CART, EN, GBM, GLM, LASSO, RF,} and \textit{RIDGE}. The depth function is therefore a description of the distribution of all possible posets based on the observed posets. Applying the ufg depth to the posets obtained from the data sets provided by the UCI repository, we get that the maximum depth value of the observed and all possible posets is $0.32$. The poset corresponding to the maximum depth value is given by the left side of Figure~\ref{fig:uci_max_bt}. The minimum depth value is zero since any poset that states that CART dominates GBM or RIDGE has to have a depth value of zero according to Theorem~\ref{cor: Dn and the sample}. 

In order to compare our approach with other existing approaches, we need to select a representative poset. Both the extended Bradley-Terry model, see~\cite{davidson70}, and the generalized stochastic dominance approach, see~\cite{jansen23} provide (partial) performance order structures that have the most evidence from their respective point of view. For comparison, we choose the poset that corresponds to the highest depth value, as it contains the structure that is most supported by the observed posets. In our case, this is the poset given by Figure~\ref{fig:uci_max_bt} (left).

The original Bradley-Terry model, see~\cite{btl}, was defined for total orders only. It is based on pairwise comparisons and assumes that these pairwise comparisons are independent. There are several approaches to extending the Bradley-Terry model, e.g.~\cite{davidson70, rao67}. We focus on the extension of the Bradley-Terry model developed by~\cite{davidson70}. This model was originally developed to analyze pairwise comparison data, where the participants can indicate a preference as well as have no preference. \cite{sinclair82} showed that this approach can be rewritten as a generalized linear model using the Poisson distribution with log link. Here, ties are considered to occur when there is no preference between two items. This extended Bradley-Terry model provides us with so-called worth parameters and a discrimination parameter that includes the no preferences of all comparisons. In our situation, the worth parameters model the latent performance structure of a classifier compared to another classifier. More precisely, if $\pi_i$ denotes the worth of classifier $i$ (with $\sum_{i} \pi_i = 1$) and $\nu$ the discrimination parameter, then the model assumes that the probability that classifier $i$ is preferred over classifier $j$ is given by $\pi_i / (\pi_i + \pi_j + \nu \sqrt{\pi_i\pi_j})$. Therefore, the classifier with the highest worth parameter is assumed to dominate all other classifiers in a pairwise comparison. Since the worth parameters are values in $[0,1]$, it is unlikely that two classifiers will have exactly the same worth, and thus in most cases a total order results. 

Note that Davidson developed this extension of the Bradley-Terry method with the aim of including Luce's choice axiom, see~\cite{chambers16}(Chapter 7). This means that the extended Bradley-Terry model assumes that the choice of one item over another is not influenced by other items. Furthermore, for two fixed classifiers $i$ and $j$, it is assumed for the estimation of the worth parameters that all observations discussing the preference of these two classifiers are independent of each other. Thus, pairwise comparisons are again assumed to be independent. However, this seems questionable because, for a fixed data set, the pairwise comparisons between the performance of classifiers cannot be guaranteed to be independent. In other words, comparing the performance of classifiers $i$ and $j$ is often related to comparing the performance of classifiers $i$ and $k$ on a fixed data set. This suggests that the extended Bradley-Terry model may not be appropriate in this situation of comparing classifiers. In general, it is questionable whether the pairwise comparisons here can be modeled by a stochastic approach for a fixed data set. In contrast, in the ufg approach, the poset-valued observation and the underlying set cannot be reduced to pairwise comparisons. Instead, the posets are considered as a whole observation. More specifically, it is necessary to reflect that the relationship between the pairwise comparisons defines a poset. %In particular, they are not assumed to be independent. 

Overall, the objective of the extended Bradley-Terry model is to estimate the true underlying worth parameters and thus obtain a total order of the items. In order to allow a proper comparison with our ufg method, we, therefore, need to select a single poset that represents our ufg depth approach. Since the ufg depth gives us a description of the distribution on all possible posets, we choose the poset corresponding to the highest ufg depth value.

%A further and related point is how we collect and use the observed posets in the extended Bradley-Terry model. For each data set, we observe a poset and to apply the extended Bradley-Terry model we use it as a set of pairwise comparisons. Thus, these observation are regarded within the ontic view in our application of the extended Bradley-Terry method.
%In contrast, in the ufg approach, the poset-valued observation and the underlying set cannot be reduced to pairwise comparisons. Instead, the posets are considered as a whole observation. More specifically, it is necessary to reflect that the relationship between the pairwise comparisons defines a poset. In particular, they are not assumed to be independent. Therefore, we choose to refer to the poset values as non-standard data rather than ontic in the context of ufg depth. 

%\textcolor{red}{@alle: ich tendiere gerade dazu, diesen Teil eher wegzulassen. In der extended Version wird auf die Bedeutung von "no preference" nicht wirklich eingegangen. Daher würde ich hier eher nur klarstellen was wir machen, aber das nicht in den Vergleich zu BT setzen. --> Dann finde ich, dass das aber eher unten rein passt. zu indistinguishable ich eigentlich noch nix gefunden im bezug auf das Bradley Terry} As pointed out in Section~\ref{sec: prop_ufg_depth}, this underlines the aspect that we see the ties that occur in the partial orders as incomparabilities that are part of the observation. In particular, we do not assume that when a tie occurs, the two items are indistinguishable. 

Let us now compare the extended Bradley-Terry model and the ufg method using the data set discussed in the subsection above. The estimation of the extended Bradley-Terry model is based on pairwise comparisons and therefore on the data shown in Figure~\ref{fig:heatmap_uci}. The estimated worth parameters are now sorted and we obtain the order shown in Figure~\ref{fig:uci_max_bt} (right). The entire estimated extended Bradley-Terry model can be found in \ref{app: BT estimation}. Comparing our approach, see Figure~\ref{fig:uci_max_bt} (left), with the extended Bradley-Terry method, we see that the poset corresponding to the extended Bradley-Terry model is a linear extension of the poset representing the highest ufg depth value. Note that for a poset corresponding to the second highest ufg depth value, the result of the extended Bradley-Terry method is no longer a linear extension. For this poset with the second highest ufg depth, the performance of GBM is below that of RF. This analysis shows that the extended Bradley-Terry method provides a stronger structure than our ufg approach. As most of the observed posets are also not a total order, this seems to support our approach as it does not impose a total order structure. On the other hand, when we need to make a decision on exactly one classifier, the extended Bradley-Terry model gives us a classifier that dominates all, rather than just a selection, even if the evidence is weak. 

%\begin{itemize}
%    \item \textcolor{red}{über indistinguishability habe ich so tatsächlich bei Bradley Terry nichts gefunden, aber ich hab auch das Luce buch nicht zur Hand. In then papern wird meist einfach über no preference gesprochen und das so definiert...}
%    \item \textcolor{red}{über ties vs incompbarability und indistighuishabel sprechen und bei dem worth parameter siehe Christophs mail 7.2  --> Luce Buch evtl direkt? Das passt vielleicht ganz gut}
%\end{itemize}

Finally, we want to use the stochastic dominance approach of~\cite{jansen23} to further clarify the difference of our approach to others. The mentioned paper analyzes, similar to the present one, the performance of classifiers with respect to several criteria on several data sets simultaneously. However, both the method and the objective differ fundamentally from those of the present paper: While we include the entire distribution of the underlying poset-valued random variables in our analysis, the authors in \cite{jansen23} compose their representative ordering on the classifiers using a binary generalized stochastic dominance (GSD) relation.

The rough idea of this GSD-based approach is to first embed the range of the multivariate performance measure in a special type of ordered set, a so-called preference system, which then allows 
 for also formalizing the entire information originating from the cardinal dimensions of the performance measure. A classifier is then judged at least as good as a competitor (similar to classic stochastic dominance) if its expected utility is at least as high with respect to every utility function representing (both the ordinal and the cardinal parts of) the preference system.\footnote{For a more detailed discussion of generalized stochastic dominance, see also~\cite{uaiall}.} Opposed to GSD, our methodology is thus particularly applicable when it is not clear how the edges within each observed partial order were obtained. Second, while the objective of the present paper is to investigate descriptively how central and outlying partial orders are over the set of classifiers under investigation (this leads to a description of the distribution on all posets), \cite{jansen23} rather investigates if dominance relations hold between classifiers over a sample/population of data sets. This is particularly noticeable in the fact that -- while our analysis takes place at the purely descriptive level -- the authors there go further and test the descriptive GSD order obtained edge by edge for statistical significance.
%In summary, we can say that the methodology presented here aims to determine which orders typically and atypically, i.e. situation-\textit{dependent}, exist between the classifiers, while the methodology in \cite{jansen23} aims to detect situation-\textit{independent} dominance relations (up to statistical uncertainty).}
%{\textcolor{cyan}{Wie gesagt, das sehe ich wirklich nur als ein Teil der Analyse und so dargestellt schmaelert es den Wert der Methode... Vorschlag: In summary, we can say that the methodology presented here can be used  to determine which orders typically and atypically, i.e....}}
Despite the clear differences in content between the methods, it is interesting to work out those aspects that can nevertheless be compared with each other.  As mentioned above, our method also offers the possibility of extracting a particularly representative partial order of the classifiers under consideration, namely one of the orders with maximum depth. For the analyzed data, this deepest order is unique and shown on the left side of Figure~\ref{fig:uci_max_bt}. If we compare this order with the descriptive GSD order from \cite{jansen23} (middle of Figure~\ref{fig:uci_max_bt}), we notice that both Hasse diagrams are very similar at first glimpse. The main difference in this case is that the analysis in \cite{jansen23} identifies a clear best classifier (GBM), while the deepest poset obtained by our method has three undominated elements (GBM, RF, and BS). The stronger structure of the GSD ordering can presumably be explained by the fact that it also exploits the cardinal information encoded in the individual quality metrics, whereas we perform a purely ordinal analysis.

Note that generally, one can not make definite statements. There are extreme situations where one method gives a very dense order as a result whereas the other method gives a very sparse poset: If the single performance measures are highly anti-correlated, then the method presented in this paper might lead to very sparse poset data and thus also to a very sparse order as the deepest data point. In contrast, the method of GSD only relies on the marginal distribution of the single performance vectors of the single classifiers over the data sets and not on the correlation structure between these vectors for different classifiers. Therefore, this method can still lead to very dense orderings as a result. On the other hand, if one has a strongly correlated structure of the performance measures and if additionally, the obtained orders are very similar on the majority of data sets, then the data depth method from here would give a very dense partial order. If at the same time, a small minority of measures is very different (i.e., outlying), then the method of \cite{jansen23} might lead to a very sparse result because it uses the cardinal scale of the performance measures.

\subsection{OpenML Repository: Demonstration of UFG Method}\label{sec: openml}
Now, we analyze data sets given by the OpenML Repository, see~ \cite{OpenML2013}, to have a detailed discussion on the performance order of different classifiers. With this, we demonstrate the richness and great variety of descriptive analysis options that are possible using our ufg method.

\subsubsection{Data Set}\label{sec:dataset}
To showcase the application of the ufg depth on machine learning algorithms we use openly available data from the OpenML benchmarking suite \cite{OpenML2013}. OpenML shares data sets and corresponding evaluations of classifiers based on different performance measures.

In our comparison we are interested in the performance of the following supervised learning methods: \textit{Random Forests} (RF, implemented in the R-package \texttt{ranger} \cite{ranger}), \textit{Decision Tree} (CART, implemented via the \texttt{rpart} library \cite{therneau2015package}),  \textit{Logistic regression} (LR), \textit{L1-penalized logistic regression} (LASSO, implemented through the \texttt{glmnet} library \cite{friedman2021package}), and \textit{k-nearest neighbors} (KNN, through the \texttt{kknn} library \cite{epub1769}).
For all methods, the choice of parameter settings depends on the goal of the user who uploaded the experiment and its results. Since different users may have different goals, this analysis does not necessarily extend to general statements about the performance of hyperparameter-tuned versions of the algorithms, where the goal is to increase their performance. 
%As stated in the OpenML experiment-documentation all methods are run with default settings of the corresponding libraries. Hence, our application analyzes the behavior of methods using default settings and does not necessarily extend to general statements about the performance of hyperparameter-tuned versions of the respective algorithms. 
The algorithms were chosen as a selection of widely used supervised learning methods that perform reasonably without much tuning, in contrast to methods such as neural networks or boosting, which require considerable tuning to perform well.

\begin{figure}
    \centering
    \begin{subfigure}{.4\textwidth}
        \centering
        \includegraphics[scale = 0.5]{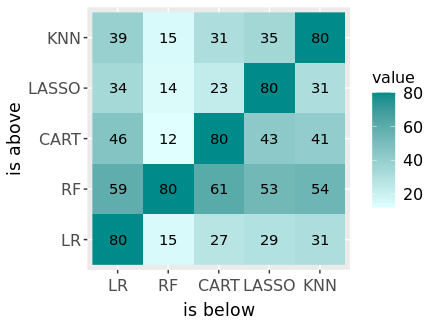}
    \end{subfigure}\hfill
    \begin{subfigure}{.4\textwidth}
        \centering
        \includegraphics[scale = 0.5]{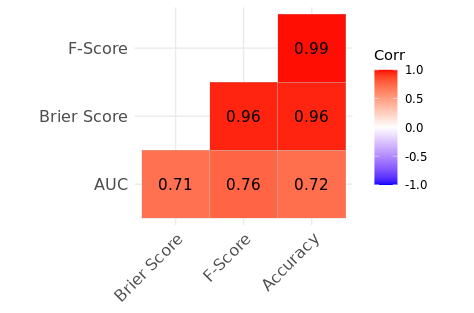}
    \end{subfigure}
    \caption{OpenML: Heatmap representing the sum-statistics, see Section~\ref{sec:properties}, based on all four performance measures (left). Compressed correlation matrix between the calculated performance measures (right). \textit{AUC} is \textit{area under the curve} for short.}
    \label{fig:heatmap}
\end{figure}

From the available data sets for which results for all the above algorithms are available in the OpenML database, we limit our analysis to binary classification data sets with more than 450 and less than 10000 observations, leading us to a total of 80 data sets for comparison. The data sets come from a variety of domains and strongly vary in their class balance as well as their overall difficulty.
Included in our multidimensional criteria comparison are the measures \textit{area under the curve}, \textit{F-score}, \textit{predictive accuracy}, and \textit{Brier score}. These performance measures capture different aspects of performance, especially in the case of unbalanced data sets. Figure~\ref{fig:heatmap} (right) shows the computed correlation between the performance measures.

The construction of the poset-valued data is analogous to the procedure described at the beginning of Section~\ref{sec:application} and in Section~\ref{sec: uci}. Since we consider here 80 data sets this leads to 80 observed posets. When including all four performance measures in the construction of the posets, we obtain that 58 of the 80 posets are unique.
%The resulting poset-valued set consists of $80$ posets, $58$ of which are unique. Each of the $58$ unique posets have a different depth value. 
The sum-statistics, see Section~\ref{sec: prop_ufg_depth}, which count for each pair the number of occurrences along the $80$ posets, can be seen in Figure~\ref{fig:heatmap} (left). It shows that RF is very often above all other methods. So if one only looks at the sum-statistics RF is clearly the strongest method. The other methods are more balanced with respect to each other. Note that due to reflexivity the diagonal is always $80$.

Finally, we want to highlight a structural observation about the order dimension of posets, which applies to all posets ordering five items. The order dimension of a finite poset is defined as the minimum number of linear extensions of this poset whose intersection is equal to this poset. In the case of posets based on five items, the order dimension is always less than or equal to two.\footnote{We established this fact by simply computing for each poset on $5$ items its order dimension, see Footnote~\ref{footn: github} for the code.} This implies that each poset can be constructed by an intersection of at most two total orders. Thus, theoretically, two different performance measures could be sufficient to obtain all possible posets. (This is indeed possible, consider for example two measures where one gives more weight to true positives and the other gives more weight to true negatives. Then any combination of two linear orders can be obtained.) In contrast, for the data sets given by the UCI repository with eight items, two performance measures are not sufficient to obtain every possible poset on eight items. For example, the crown $S_4^0$, see \cite{TROTTER197485}, is a poset on eight items with order dimension four. Therefore, in the case of Section~\ref{sec: uci}, at least four performance measures are required. Note that the above observation should not be taken as advice on how to choose the (number of) performance measures. One aspect is that, given a set of performance measures $A$, it is questionable whether there are two performance measures that represent exactly the same order as that obtained by the set of performance measures $A$. This issue of reducing the number of performance measures based on the order dimension is discussed in the next section. In any case, the decision on which and how many measures to include in an analysis should always be based on substance considerations.

%The corresponding posets then result from the multidimensional criteria comparison. It should be noted that rescaling the performance measures does not change the posets. This follows from the fact that the posets defined here do not depend on the absolute differences but on whether or not the multiple performance measures are better in all dimensions.

\subsubsection{Analysis using UFG Depth}\label{sec:analysis}
In this subsection, we give a detailed illustration of how the ufg depth gives an insight into the performance structure of classifiers. This is done using the data sets, classifiers, and performance evaluations of the OpenML Repository described in the section above.
This analysis consists of two parts: First, we discuss the poset-valued data obtained by using all four performance measures together. Here, we focus on aspects like what the deepest posets have in common as well as considerations on dispersion. In the second part, we tackle the question of how strong the influence of the choice of the performance measures on the resulting poset valued data as well as the depth function is. 

\begin{figure}
    \centering
    \includegraphics[width=.8\linewidth]{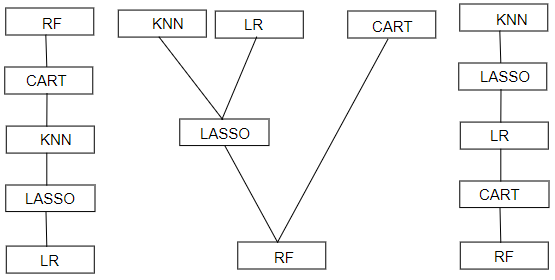}
    %\begin{subfigure}{.8\textwidth}
    %    \centering
         
    %\end{subfigure}%
    %\begin{subfigure}{.2\textwidth}
    %    \centering
    %    \includegraphics[scale = 0.5]{blocher23_part1_min_depth_all.png}
    %\end{subfigure}
    \caption{OpenML based on all four performance measures: Poset with maximal depth based on all possible posets is plotted on the left. The poset with minimal ufg depth restricted to the observed one can be seen in the middle. The poset on the right denotes the poset with minimal depth value based on all possible posets.}
    \label{fig:max_min}
\end{figure}

Let us start with discussing the analysis based on the poset valued data using all four performance measures for their construction. This gives us a set of 80 posets corresponding to the heatmap in Figure~\ref{fig:heatmap} (left). Here, 58 of the posets are unique. Based on these 80 observed posets we compute the empirical ufg depth. Evaluating this empirical ufg depth over the entire set of possible posets $\Pcal$, we find that each of the 4231 posets has a unique depth value.
The most central poset based on all possible posets with maximum depth value is a total order and can be seen in Figure~\ref{fig:max_min} (left). This poset is also observed and has a depth value of 0.34. Note that the poset with the highest depth value also has the most duplicates, meaning it is the most common pattern given by the posets obtained by the data sets. As described in Section~\ref{sec:comparison}, we are interested in the distribution of the observed posets. Nevertheless, we can consider the poset with the highest depth value as the poset whose structure is the most common one. Or, in other words, this poset is the one that is most supported based on all observations. 
%Comparing this to Figure~\ref{fig:heatmap} or, e.g., the results in~\cite{jnsa2022}, we see many similarities, such as LR often has worse performance than the other algorithms, and RF dominates all other algorithms in many cases. In contrast to the sum-statistics which here give a representative poset, the strength of our method is that we not only obtain one single poset structure, but also a distribution over the set of posets. Note that in general the order given by the sum-statistics is not a poset, i.e. their might exist cycles.

%While the poset with the highest depth value shows a clear dominance structure for CART, LASSO and KNN, this is not clearly stated by the sum-statistics. 

\begin{figure}
    \centering
    \includegraphics[scale = 0.6]{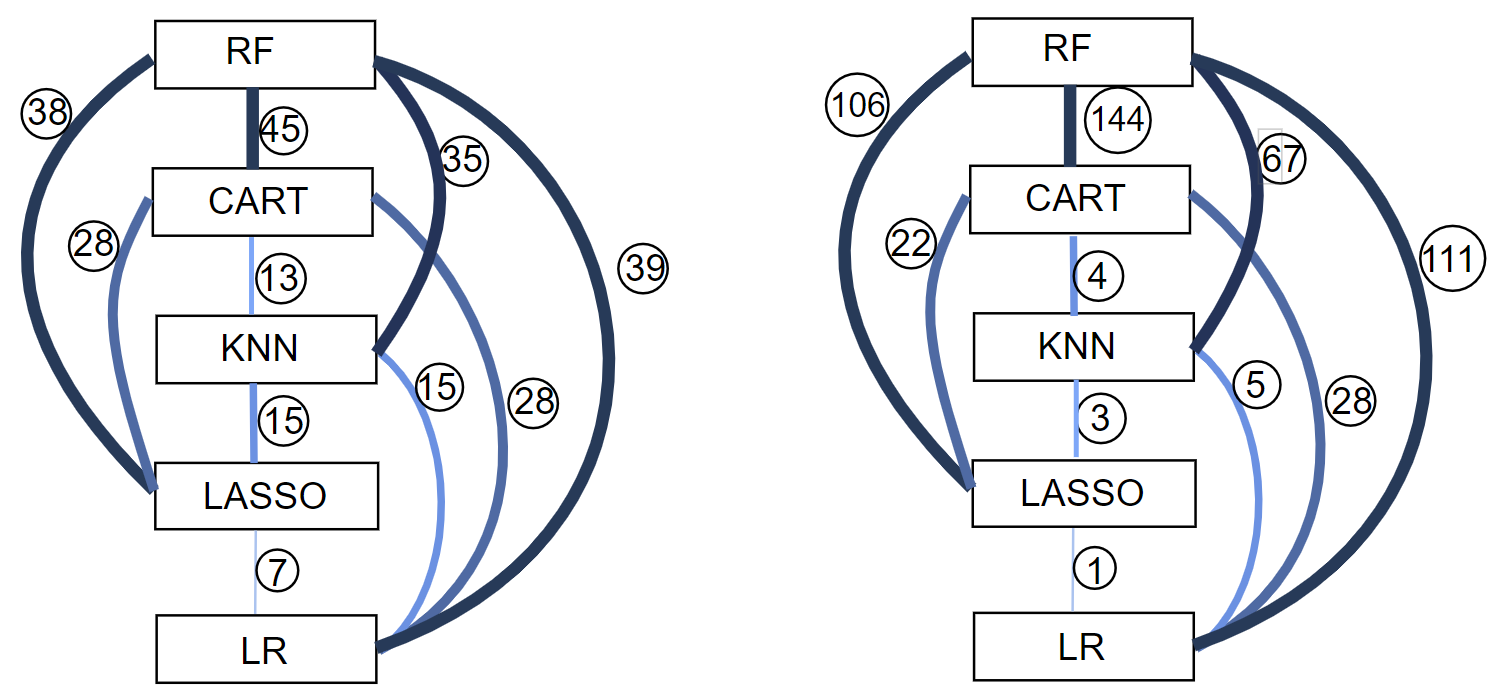}
    \caption{OpenML based on all four performance measures: Represents what the (observed) posets with the $k$ highest depth values have in common. On the left-hand side, we restrict the analysis to the observed posets and on the right-hand side, we focus on all possible posets. Compare with Figure~\ref{fig:max_min}, where the poset with the highest depth value is plotted. Here each edge number $k\in \mathbb{N}$ indicates that the $k$ deepest posets all contain this relation, but this is not true for the $k+1$ deepest poset.}
    \label{fig:Schnitt}
\end{figure}

Figure \ref{fig:Schnitt} describes which edges the posets with the $k \in \mathbb{N}$ highest depth values have in common. On the left-hand side, we focus on the observed posets and, the right-hand side is based on all possible posets. Note that while the underlying space on the right-hand side is larger, we include duplicates on the left-hand side. Thus, on the right-hand side, we can see that the deepest poset has seven duplicates in the data set. First, observe that the structure based on restricting to the observed posets or using all possible posets is very similar, i.e. the dominance structure eliminated differs only slightly between all or only the observed posets. For example, one can see that the dominance of RF over all classifiers based on all four performance measures holds for the 35 observed posets with the highest depth values and based on all possible posets for the 67 poset with the highest depth values. In particular, any other classifier dominance (like CART outperforms KNN according to all performance measures), does not hold for so many posets with the highest depth values. Note that the observed posets with the highest 46 depth values have nothing more in common. For all possible posets, this is true for the 145 posets with the highest depth value. 

Conversely, it is of interest to see what non-edges the posets have in common. Since the poset with the highest depth value is a total order, this is immediately apparent in Figure~\ref{fig:Schnitt}. The posets with $k \in \mathbb{N}$ highest depth values have those non-edges in common, which are given by the inverse ordered poset of highest depth value intersecting with the inversely already deleted ones. For example, the observed posets with the nine highest depth values have in common that the RF is not dominated by CART, CART not by KNN, and KNN not by LASSO, but they do not agree on LASSO being not dominated by LR since the posets with the observed 8th highest depth value do not agree on this.

Unlike the posets with the highest depth values, the posets with low depth values do not have much in common. The posets (both observed and not observed) only agree on RF being dominated by another classifier. After that, no structure holds. All of these posets can be seen as outliers, or in other words, the corresponding data sets produce a performance structure on the classifiers which differ from the structure given by other data sets. The observed poset with the smallest depth value, which is 0.05, is plotted in the middle of Figure~\ref{fig:max_min}. The poset on the right-hand side of Figure~\ref{fig:max_min} shows the poset corresponding to the smallest depth value (which is 0.01) based on all possible posets. 

Finally, we want to give a notion of dispersion of the depth function. Therefore, we compute the depth function for every poset $p \in \Pcal$ and compute the proportion of posets that lie in $\alpha \in [0,1]$ deepest observed depth values. For $\alpha = 0.25, 0.5$ and $0.75$ we get $0.02, 0.10$ and $0.26$. Thus, the empirical ufg depth seems to be clustered on small parts of the set $\Pcal$. Note that this impression is a little vague, one computes the proportions of posets compared to the set of all posets. In certain situations, such as for the data sets provided by the UCI repository, not all posts can be obtained because the order dimension is four, but only two performance measures are used. Therefore it may be more natural to compute the proportions in comparison to the number of posets that could be obtained at all. However, as can be seen from the discussion of the order dimension from above (Section~\ref{sec: openml}), for the case of the OpenML data set, we are not in such a situation.
Finally, it may be still more adequate to compare values for the dispersion only directly between different data sets with the same or similar parameters like the number of items.
%But this is the typical situation of application, anyhow.}}\textcolor{red}{@Georg: hier klarer Bezug auf Ordnungsdimension}

The above analysis discusses posets that represent the performance strength of classifiers using all four performance measures. The discussion of the order dimension shows that, in principle, two measures might be sufficient to define all possible posets representing the performance strength of the five classifiers. However, if the two measures both aim to represent the performance strength of the algorithms, it is questionable whether such two measures exist. For example, in the analysis done above, all four performance measures are highly correlated, see Figure~\ref{fig:heatmap} (right). It is also reasonable that the two measures that give all possible posets are derived from functions that have nothing to do with the performance of classifiers. Nevertheless, we want to discuss a heuristic approach to check whether a subset of size two of the performance measures discussed above is sufficient to obtain the same posets or at least a similar depth function. This heuristic approach is based on the idea that two performance measures with a low correlation tend to measure strength differently. Thus, using two performance measures with low correlation may give us similar posets and depth function structure as using all four performance measures at once. Therefore, we start with two performance measures that have a low correlation. To evaluate this heuristic approach, we also discuss the posets and resulting depth functions obtained by two performance measures with higher correlation.
%In particular, since it could be that the correlation of the performance measures depends on the data set structure, like the number of observations, dimension or class (im-)balance. Thus, it is not necessarily clear which performance measure can be deleted.
\begin{figure}
    \centering
    \begin{subfigure}{.5\textwidth}
        \centering
         \includegraphics[scale = 0.45]{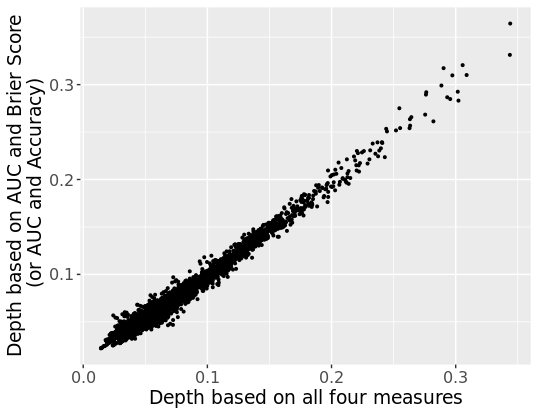}
    \end{subfigure}%
    \begin{subfigure}{.5\textwidth}
        \centering
        \includegraphics[scale = 0.45]{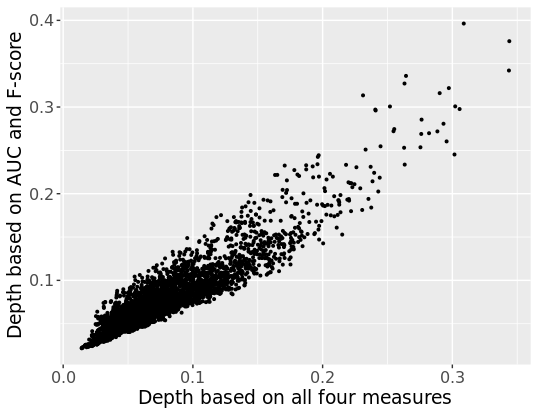}
    \end{subfigure} \\[1ex]
    %\begin{subfigure}{\textwidth}
    %    \centering
    %    \includegraphics[scale = 0.35]{part2b_all_scatter_plot_31_33.pdf}
    %\end{subfigure}
    \caption{OpenML: Each point in the scatter plot describes the depth of one poset based on the observed posets given by two different performance sets but based on the same data sets. The scatter plot on the right upper corner compares the depth of the posets based on the observed ones using \textit{area under the curve} and \textit{F score} with the observed posets using all four performance measures. }
    \label{fig:performance_correlation_scatterplot}
\end{figure}
%Let us start by considering those posets that come from two performance measures that have a low correlation. 

Based on Figure~\ref{fig:heatmap} (right), \textit{area under the curve} and \textit{Brier score} have a low correlation, as well as \textit{area under the curve} and \textit{predictive accuracy}. It is noteworthy that these two sets of performance measures give exactly the same observed posets. Therefore, the resulting depth functions are also the same. Compared to the posets obtained by using all four performance measures simultaneously, we see that four data sets produce different posets. Figure~\ref{fig:performance_correlation_scatterplot} (left) compares the depth of all possible posets based on the posets obtained by all four performance measures (x-axis) and based on the posets obtained by the \textit{area under the curve} and \textit{Brier score} measures (y-axis). Each point in the scatter plot represents a poset and the axes denote the depth values. We observe that the depth values for all possible posets seem not to differ strongly between the two construction processes using \textit{area under the curve} and \textit{Brier score} or all four performance measures. If we look at the order of the posets using the two depth functions, we can see that the orders of the posets based on the depth functions are different after all. For a single depth function, we can order the posets according to their respective depth values. Applying this to both depth functions, we find that the maximum rank shift is 1679. More specifically, the poset with the maximal rank shift has the 71st smallest depth value based on all four performance measures, and using the posets obtained by \textit{area under the curve} and \textit{Brier score} only it has the 1750th smallest depth value. In total, 50\% of the posets have a rank shift of at least 205 when comparing these two sets of performance measures. Note that this difference in the ranking is within the posets which are not observed. When restricting our analysis to the 76 posets which are observed by both sets of performance measures, then the shift is at most 6 ranks.

Now let us take the next step and look at the posets based on \textit{area under the curve} and \textit{F score}. Recalling Figure~\ref{fig:heatmap} (right), we see that these two measures still have a low correlation of 0.76 compared to all other computed correlations. Nevertheless, there are now 19 data sets that give different poset structures using either \textit{area under the curve} and \textit{F score} or all four performance measures together. This results in a greater difference in the depth values, see Figure~\ref{fig:performance_correlation_scatterplot} (right).  This is also reflected in the ranking of the posets based on the depth values. Now 50\% of the posets have a rank difference of at least 305 and the highest rank difference is 2260. Interestingly, if we look at the depth functions given by posets obtained by \textit{F~score} and \textit{predicitive accuracy}, we have a smaller shift of the rank structure compared to the depth using all four performance measures. Here the shift of 50\% of the posets is at least 226 and at most 1802. This is particularly interesting as \textit{F score} and \textit{predicitive accuracy} have the highest correlation with 0.99, see Figure~\ref{fig:heatmap} (right).

Comparing all possible subsets of size two of the above performance measures, we see that using the two performance measures with the lowest correlation does indeed lead to posets and a depth function that are most similar to those obtained by all four performance measures. However, we have seen that the rank structure of the resulting depth functions is still different from that when all four performance measures are considered simultaneously. Furthermore, using \textit{area under the curve} and \textit{F~score}, which also have a comparatively low correlation, leads to different posets and different depth functions. This does not support our heuristic approach. In particular, using the two measures with the highest correlation leads to posets and a depth function that are more similar to those using all four measures than using  \textit{area under the curve} and \textit{F~score}. Overall, this analysis highlights the point that even if two measures are theoretically sufficient to obtain all possible posets, we cannot simply pick two performance measures and expect that this will lead to all posets being obtained by more than two performance measures. This supports our idea of using multiple performance measures simultaneously. Especially as the order dimension increases with a growing number of classifiers.

\section{Conclusion} \label{sec:conclusion}

%{\textcolor{cyan}{GS:bisher nur hierhin verschoben, noch zu uebrarbeiten:
%Another aspect that we did not account for in this paper is the quantification of statistical uncertainty to reflect the fact that the performance measures are only measures of the true out of sample performances. Beyond the general need of uncertainty quantification, due to varying complexities (especially because some classifiers have many hyperparameters and others not), different classifiers (despite cross-validation, cf., e.g., \cite{varma2006bias, JMLR:v11:cawley10a}) do actually vary very much in the dispersion and bias of their performance estimates. In principle, such heterogeneous statistical uncertainty can be incorporate within our approach by building confidence intervals and by comparing the obtained intervals with/as ??? interval orders}}

%which could be taken into account for example by comparing not point estimates but confidence intervals and by comparing such intervals with interval orders.
%unsicherheit quantifizieren 

In this paper, we have shown how samples of poset-valued random variables can be analyzed (descriptively) by utilizing a generalized concept of data depth. For this purpose, we first introduced an adaptation of the simplicial depth, the so-called ufg depth, and studied some of its properties. Finally, we illustrated our framework with two examples of comparing classifiers using multiple performance measures simultaneously. In the process, we demonstrated how our approach differs from other methods used to analyze poset-valued data and highlighted the various ways in which poset-valued data can be analyzed based on the ufg depth. There are several promising avenues for future research, that include (but are not limited to):

\textbf{Other ML Problems:} Here, we focused on the comparison of classifiers. For example, the performance of different optimization algorithms could be also of interest. In this situation, one can take advantage of the fact that it is possible to tie back to the optimization problems that produced the performance structure. Thus, instead of analyzing the poset with the highest/lowest depth value, one can analyze the optimization problems that produce these posets.

\textbf{Other Performance Criteria:} Instead of using a set of unidimensional performance criteria, the analysis of classifiers with respect to other criteria could be an interesting modification. For example, one could use ROC curves, which can be easily incorporated into our order-based approach.

%\textbf{Discussion on computation time:} In Section~\ref{subsec:implementation} we briefly discussed the computation time and the difficulty of predicting it. For a deeper understanding further analyses, e.g. in form of a simulation study, would be helpful.

\textbf{Inference:} A first step towards inference for poset-valued random variables is already made by the consistency property in Section~\ref{sec:properties}. Natural next tie-in points are provided by regression and statistical testing. Together with the results for modeling in~\cite{bsj2022}, a complete statistical analysis framework for poset-valued random variables would then be achieved.

\textbf{Statistical Uncertainty:} Performance measures are only estimates of the true out-of-sample performance. Currently, we are not quantifying this underlying statistical uncertainty. Beyond the general need for uncertainty quantification, different classifiers (despite cross-validation, see e.g. \cite{varma2006bias, JMLR:v11:cawley10a}) actually differ very much in the dispersion and bias of their performance estimates. This is due to differences in the complexity of the classifiers (mainly because some classifiers have many hyperparameters and others do not).  In principle, such heterogeneous statistical uncertainty can be incorporated into our approach by constructing confidence intervals and comparing the resulting intervals with interval orders.

\textbf{Other Types of Non-Standard Data:} Our analysis framework is by no means limited to poset-valued random variables. Since the ufg depth is based on a closure operator, all non-standard data types for which a meaningful closure operator exists can be analyzed with it. As seen in \cite{bsj2022} such closure operators are easily obtained by formal concept analysis, thus, there exists a natural generalization of the ufg depth for non-standard data.
\appendix
\section{Computation Time and Complexity}\label{appendix: computation time}
Here, we state the computation time and complexity of the upper analyzed ufg depth functions. 

\subsection{UCI Repository}
We consider 16 different posets $p_1, \ldots, p_{16}$, ordering eight elements. By the definition of empirical ufg depth, see Definition~\ref{def: empirical ufg depth}, we get that we only need to consider those union-free generic sets $S$ which are a subset of $\{p_1, \ldots,$ $ p_{16}\}$. Applying Theorem~\ref{th: S+card 2 and connect} Part 1. we obtain that it is sufficient to check all subsets of $\{p_1, \ldots, p_{16}\}$ with size greater than two and test if they are union-free and generic. So we checked 65519 subsets and got that 4010 of them are union-free and generic. Testing all $S \subseteq \{p_1, \ldots, p_{16}\}$ to see if they are union-free and generic and calculating the depth of the observed 16 posets required about 2 minutes. Since computing all possible posets for eight items is not feasible in a reasonable time, we used the binary linear programming approach described in Section~\ref{subsec:implementation}. Defining the binary linear program required about 6 seconds. Calculating the highest depth value along with the corresponding poset required about 15 seconds. Calculation of the 20 highest depth values together with the corresponding posets required about 1 minute.

\subsection{OpenML Repository}
Here, we considered in each computation 80 posets which order five items. Hereby, we considered different performance measures and therefore obtained different poset sets. In what follows, we break down the computation time and complexity based on all different observed poset sets.
\vspace{0.5em}

\noindent\textit{Part 1: Using all four performance measures}

Here we see that 58 of the 80 posets $\{p_1, \ldots, p_{80}\}$ are unique (w.l.o.g. the first 58 posets are unique). We can use Theorem~\ref{th: S+card 2 and connect} to get that for all $S \in \Sscr_{obs}$ we have $2 \leq \# S \leq 8$. Still, checking every single subset $S \in \{p_1, \ldots, p_{58}\}$ with size between 2 and 8 is very time consuming, as these are 2262985652 sets (ignoring duplicated posets). Therefore we use the connectedness property, see Theorem~\ref{th: S+card 2 and connect}. With this, computing all union-free generic sets based on the unique posets $\{p_1, \ldots, p_{58}\}$ took a total of about 5 hours.
We obtained 159382 union-free generic sets. Based on this, together with a weighting according to the duplicates, the computation time of the ufg depth for all possible posets is about 37 minutes.
\vspace{0.5em}

\noindent\textit{Part 2: Using area under the curve and Brier score (or area under the curve and predictive accuracy)}

Here we see that 57 of the 80 posets $\{p_1, \ldots, p_{80}\}$ are unique (w.l.o.g. the first 57 posets are unique). We can use Theorem~\ref{th: S+card 2 and connect} to get that for all $S \in \Sscr_{obs}$ we have $2 \leq \# S \leq 8$. Still, checking every single subset $S \in \{p_1, \ldots, p_{57}\}$ with size between 2 and 8 is very time consuming, as these are 1957698535        sets (ignoring duplicated posets). Therefore we use the connectedness property, see Theorem~\ref{th: S+card 2 and connect}. With this, computing all union-free generic sets based on the unique posets $\{p_1, \ldots, p_{57}\}$ took a total of about 3 hours.
We obtained 140118 union-free generic sets. Based on this, together with a weighting according to the duplicates, the computation time of the ufg depth for all possible posets is about 9 minutes.
\vspace{0.5em}

\noindent\textit{Part 3: Using area under the curve and F-score}

Here we see that 52 of the 80 posets $\{p_1, \ldots, p_{80}\}$ are unique (w.l.o.g. the first 52 posets are unique). We can use Theorem~\ref{th: S+card 2 and connect} to get that for all $S \in \Sscr_{obs}$ we have $2 \le \# S \le 7$. Still, checking every single subset $S \in \{p_1, \ldots, p_{52}\}$ with size between 2 and 7 is very time consuming, as these are 157036191 sets (ignoring duplicated posets). Therefore we use the connectedness property, see Theorem~\ref{th: S+card 2 and connect}. With this, computing all union-free generic sets of $\{p_1, \ldots, p_{52}\}$ took a total of about 36 minutes.
We obtained 69641 union-free generic sets. Based on this, together with a weighting according to the duplicates, the computation time of the ufg depth for all possible posets is about 5 minutes.
\vspace{0.5em}

\noindent\textit{Part 4: Using predictive accuracy and F-score}

Here we see that 56 of the 80 posets $(p_1, \ldots, p_{80})$ are unique (w.l.o.g. the first 56 posets are unique). We can use Theorem~\ref{th: S+card 2 and connect} to get that for all $S \in \Sscr_{obs}$ we have $2 \le \# S \le 8$. Still, checking every single subset $S \in \{p_1, \ldots, p_{56}\}$ with size between 2 and 8 is very time consuming, as these are 1689096277 sets (ignoring duplicated posets). Therefore we use the connectedness property, see Theorem~\ref{th: S+card 2 and connect}. With this, computing all union-free generic sets of $\{p_1, \ldots, p_{56}\}$ took a total of about 2 hours.
We obtained 120145 union-free generic sets. Based on this, together with a weighting according to the duplicates, the computation time of the ufg depth for all possible posets is about 7 minutes. 

\section{Estimated Extended Bradley Terry Model on the Posets Derived by the UCI Repository}\label{app: BT estimation}

In this section, we briefly present the evaluation of the data sets provided by the UCI repository using the extended Bradley-Terry model, see~\cite{btl, davidson70}. It is based on the discussion and introduction in Section~\ref{sec: uci}. 
%For further reading we refer to~\cite{btl, sinclair82, davidson70}. 

\cite{sinclair82} showed that the extended Bradley-Terry model can be rewritten as a generalized linear model. More precisely, for two different classifiers $i$ and $j$, let $m_{i>j}$ be the number of comparisons that classifier $i$ is preferred over classifier $j$. The parameters $\pi_i$ and $\pi_j$ are the worth parameters of the respective class (note that over all classifiers we assume that $\sum_{\ell} \pi_{\ell} = 1$), and $\nu$ is the discrimination parameter, which indicates the tendency of an answer to be no preference.
The extended Bradley-Terry model is then a generalized linear model with Poisson distribution and log link. The linear predictor is given by
\begin{align*}
    &\log(m_{i>j}) = \mu_{ij} + \frac{1}{2}\log(\pi_i) - \frac{1}{2}\log(\pi_j),\quad \text{ and }\\
    &\log(m_{i\sim j}) = \mu_{ij} + \log(\nu)
\end{align*}
with $\mu_{ij} = \ln{m} - \ln{\left(\sqrt{\pi_i / \pi_j} + \sqrt{\pi_j / \pi_i}\right)}$ where $m$ is the total number of pairwise comparisons.

We applied this to the posets provided by the UCI repository (this code is written in R):
\begin{lstlisting}
> result_Uci_glm <- glm(cum_sum ~ -1 + mu + undecided +
+                       BS + CART + EN + GBM + GLM +
+                       LASSO + RF +RIDGE,
+                       family = 'poisson', 
+                       data = design_mat) 
                        # loglink is default
> summary(result_Uci_glm)

Call:
glm(formula = cum_sum ~ -1 + mu + undecided + BS + CART +
    EN + GBM + GLM + LASSO + RF + RIDGE, 
    family = "poisson", data = design_mat)

Coefficients: (1 not defined because of singularities)
              Estimate Std. Error z value Pr(>|z|)    
muBS_CART      0.73676    0.30950   2.380 0.017292 *  
muBS_EN        1.41555    0.26296   5.383 7.32e-08 ***
muBS_GBM       1.52960    0.25459   6.008 1.88e-09 ***
muBS_GLM       1.52983    0.25457   6.010 1.86e-09 ***
muBS_LASSO     1.39271    0.26458   5.264 1.41e-07 ***
muBS_RF        1.53411    0.25424   6.034 1.60e-09 ***
muBS_RIDGE     1.44596    0.26078   5.545 2.94e-08 ***
muCART_EN      1.17764    0.28372   4.151 3.32e-05 ***
muCART_GBM     0.63859    0.31463   2.030 0.042392 *  
muCART_GLM     0.82902    0.30456   2.722 0.006489 ** 
muCART_LASSO   1.21233    0.28135   4.309 1.64e-05 ***
muCART_RF      0.72483    0.31013   2.337 0.019432 *  
muCART_RIDGE   1.12363    0.28728   3.911 9.18e-05 ***
muEN_GBM       1.36676    0.26644   5.130 2.90e-07 ***
muEN_GLM       1.45551    0.26007   5.597 2.19e-08 ***
muEN_LASSO     1.53309    0.25432   6.028 1.66e-09 ***
muEN_RF        1.40995    0.26336   5.354 8.62e-08 ***
muEN_RIDGE     1.53178    0.25441   6.021 1.74e-09 ***
muGBM_GLM      1.51639    0.25562   5.932 2.99e-09 ***
muGBM_LASSO    1.34024    0.26826   4.996 5.85e-07 ***
muGBM_RF       1.53066    0.25451   6.014 1.81e-09 ***
muGBM_RIDGE    1.40282    0.26392   5.315 1.06e-07 ***
muGLM_LASSO    1.43646    0.26145   5.494 3.93e-08 ***
muGLM_RF       1.52866    0.25466   6.003 1.94e-09 ***
muGLM_RIDGE    1.48011    0.25827   5.731 1.00e-08 ***
muLASSO_RF     1.38664    0.26500   5.233 1.67e-07 ***
muLASSO_RIDGE  1.52747    0.25475   5.996 2.02e-09 ***
muRF_RIDGE     1.44107    0.26114   5.518 3.42e-08 ***
undecided      0.37166    0.10989   3.382 0.000720 ***
BS             0.55687    0.17803   3.128 0.001760 ** 
CART          -1.24203    0.21482  -5.782 7.40e-09 ***
EN            -0.09093    0.17423  -0.522 0.601721    
GBM            0.68258    0.18096   3.772 0.000162 ***
GLM            0.43440    0.17586   2.470 0.013505 *  
LASSO         -0.15226    0.17485  -0.871 0.383885    
RF             0.57238    0.17835   3.209 0.001331 ** 
RIDGE               NA         NA      NA       NA    
---
Signif. codes:  0 '***' 0.001 '**' 0.01 '*' 0.05  
                '.' 0.1 ' ' 1

(Dispersion parameter for poisson family taken to be 1)

Null deviance: 975.153  on 84  degrees of freedom
Residual deviance:  61.312  on 48  degrees of freedom
AIC: 404.11

Number of Fisher Scoring iterations: 5
\end{lstlisting}

Note that RIDGE is the reference level. Thus, the estimate of the worth parameter is the logarithm of zero. The upper result now gives us the estimated probability that classifier $i$ is preferred over classifier $j$. For example, consider the estimated probability that CART will be outperformed by GBM. Note that this is the dominance structure that exists for the largest number $k\in \{1, \ldots, 16\}$ in $\cap_{p \in \{p_{(1)}, \ldots, p_{(k)}\}}\: p$, where $p_{(1)}, \ldots, p_{(16)}$ are the observed posets, ordered in descending order of their ufg depth value. We have that this preference has a probability of $0.81$. The calculation is as follows: The estimated worth parameters are given by $\pi_{\text{GBM}} = \exp(2\cdot 0.68258) \approx 3.92$ and $\pi_{\text{CART}} = \exp(2 \cdot (-1.24203)) \approx 0.08$. With the estimated discrimination parameter $\nu = \exp(0.37166) \approx 1.45$ we get the estimated probability by $\pi_{\text{GBM}} / (\pi_{\text{GBM}} + \pi_{\text{CART}} + \nu \sqrt{\pi_{\text{GBM}}\pi_{\text{CART}}}) \approx 0.81$. The estimated probability that there is no preference between GBM and CART is $\nu \sqrt{\pi_{\text{GBM}}\pi_{\text{CART}}} / (\pi_{\text{GBM}} + \pi_{\text{CART}} + \nu \sqrt{\pi_{\text{GBM}}\pi_{\text{CART}}}) \approx 0.17.$

\section*{Acknowledgments}
We sincerely thank all four anonymous reviewers of the ISIPTA conference 2023 in Oviedo, Spain for valuable comments that helped to improve the paper. Moreover, we want to thank all participants at the ISIPTA'23 conference in Oviedo for all very helpful discussions. We are also grateful to the two reviewers for their review of this extended version. Hannah Blocher and Georg Schollmeyer gratefully acknowledge the financial and general support of the LMU Mentoring program. Hannah Blocher sincerely thanks Evangelisches Studienwerk Villigst e.V. for funding and supporting.

\section*{Competing Interest}
There exist no competing interests.

\section*{Author Contributions}
CRediT roles:
\begin{itemize}
    \item \textit{Conceptualization}: Hannah Blocher (lead); Georg Schollmeyer (supporting); Christoph Jansen (supporting)
    \item \textit{Methodology}: Hannah Blocher (lead); Georg Schollmeyer (supporting);  Christoph Jansen (supporting)
    \item \textit{Software}: Hannah Blocher (lead); Georg Schollmeyer (supporting)
    \item \textit{Validation}: Hannah Blocher (lead);  Georg Schollmeyer (supporting)
    \item \textit{Formal Analysis}:  Hannah Blocher (lead); Georg Schollmeyer (supporting); Christoph Jansen (supporting)
    \item \textit{Data Curation}: Malte Nalenz (lead); Hannah Blocher (supporting)
    \item \textit{Writing - Original Draft}: Hannah Blocher (lead);  Georg Schollmeyer (supporting); Christoph Jansen (supporting); Malte Nalenz (supporting) 
    \item \textit{Writing - Review \& Editing}: Hannah Blocher (equal);  Georg Schollmeyer (equal); Christoph Jansen (equal)
    \item \textit{Visulaization}: Hannah Blocher
    \item \textit{Supervision}: Hannah Blocher
    \item \textit{Project Administration}: Hannah Blocher 
\end{itemize}

\section*{Fundings}
 Hannah Blocher and Georg Schollmeyer are financially and generally supported by the LMU Mentoring Program. Hannah Blocher's dissertation project is supported by a scholarship of the Evangelisches Studienwerk Villigst e.V.

%\section{}
%\label{}



\bibliographystyle{elsarticle-harv} 
\bibliography{blocher23.bib}

\begin{thebibliography}{64}
\expandafter\ifx\csname natexlab\endcsname\relax\def\natexlab#1{#1}\fi
\providecommand{\url}[1]{\texttt{#1}}
\providecommand{\href}[2]{#2}
\providecommand{\path}[1]{#1}
\providecommand{\DOIprefix}{doi:}
\providecommand{\ArXivprefix}{arXiv:}
\providecommand{\URLprefix}{URL: }
\providecommand{\Pubmedprefix}{pmid:}
\providecommand{\doi}[1]{\href{http://dx.doi.org/#1}{\path{#1}}}
\providecommand{\Pubmed}[1]{\href{pmid:#1}{\path{#1}}}
\providecommand{\bibinfo}[2]{#2}
\ifx\xfnm\relax \def\xfnm[#1]{\unskip,\space#1}\fi
%Type = Article
\bibitem[{Armstrong(1974)}]{armstrong74}
\bibinfo{author}{Armstrong, W.}, \bibinfo{year}{1974}.
\newblock \bibinfo{title}{Dependency structures of data base relationships}.
\newblock \bibinfo{journal}{International Federation for Information Processing Congress} \bibinfo{volume}{74}, \bibinfo{pages}{580--583}.
%Type = Article
\bibitem[{Baker and Scarf(2021)}]{baker21}
\bibinfo{author}{Baker, R.}, \bibinfo{author}{Scarf, P.}, \bibinfo{year}{2021}.
\newblock \bibinfo{title}{Modifying {B}radley--{T}erry and other ranking models to allow ties}.
\newblock \bibinfo{journal}{IMA Journal of Management Mathematics} \bibinfo{volume}{32}, \bibinfo{pages}{451--463}.
%Type = Inproceedings
\bibitem[{Bastide et~al.(2000)Bastide, Pasquier, Taouil, Stumme and Lakhal}]{bastide00}
\bibinfo{author}{Bastide, Y.}, \bibinfo{author}{Pasquier, N.}, \bibinfo{author}{Taouil, R.}, \bibinfo{author}{Stumme, G.}, \bibinfo{author}{Lakhal, L.}, \bibinfo{year}{2000}.
\newblock \bibinfo{title}{Mining minimal non-redundant association rules using frequent closed itemsets}, in: \bibinfo{editor}{Lloyd, J.}, \bibinfo{editor}{Dahl, V.}, \bibinfo{editor}{Furbach, U.}, \bibinfo{editor}{Kerber, M.}, \bibinfo{editor}{Lau, K.}, \bibinfo{editor}{Palamidessi, C.}, \bibinfo{editor}{Pereira, L.}, \bibinfo{editor}{Sagiv, Y.}, \bibinfo{editor}{Stuckey, P.} (Eds.), \bibinfo{booktitle}{Computational Logic --- CL 2000}, \bibinfo{publisher}{Springer}. pp. \bibinfo{pages}{972--986}.
%Type = Article
\bibitem[{Benavoli et~al.(2016)Benavoli, Corani and Mangili}]{bcm2016}
\bibinfo{author}{Benavoli, A.}, \bibinfo{author}{Corani, G.}, \bibinfo{author}{Mangili, F.}, \bibinfo{year}{2016}.
\newblock \bibinfo{title}{Should we really use post-hoc tests based on mean-ranks?}
\newblock \bibinfo{journal}{The Journal of Machine Learning Research} \bibinfo{volume}{17}, \bibinfo{pages}{152--161}.
%Type = Article
\bibitem[{Bertet et~al.(2018)Bertet, Demko, Viaud and Gu{\'e}rin}]{bertet18}
\bibinfo{author}{Bertet, K.}, \bibinfo{author}{Demko, C.}, \bibinfo{author}{Viaud, J.}, \bibinfo{author}{Gu{\'e}rin, C.}, \bibinfo{year}{2018}.
\newblock \bibinfo{title}{Lattices, closures systems and implication bases: A survey of structural aspects and algorithms}.
\newblock \bibinfo{journal}{Theoretical Computer Science} \bibinfo{volume}{743}, \bibinfo{pages}{93--109}.
%Type = Misc
\bibitem[{Blocher and Schollmeyer(2023)}]{blocher23b}
\bibinfo{author}{Blocher, H.}, \bibinfo{author}{Schollmeyer, G.}, \bibinfo{year}{2023}.
\newblock \bibinfo{title}{Data depth functions for non-standard data by use of formal concept analysis}.
\newblock \URLprefix \url{https://www.foundstat.statistik.uni-muenchen.de/personen/mitglieder/blocher/blocheretal_properties23.pdf}. \bibinfo{note}{[Accessed: 21.11.2023]}.
%Type = Inproceedings
\bibitem[{Blocher et~al.(2022)Blocher, Schollmeyer and Jansen}]{bsj2022}
\bibinfo{author}{Blocher, H.}, \bibinfo{author}{Schollmeyer, G.}, \bibinfo{author}{Jansen, C.}, \bibinfo{year}{2022}.
\newblock \bibinfo{title}{Statistical models for partial orders based on data depth and formal concept analysis}, in: \bibinfo{editor}{Ciucci, D.}, \bibinfo{editor}{Couso, I.}, \bibinfo{editor}{Medina, J.}, \bibinfo{editor}{Slezak, D.}, \bibinfo{editor}{Petturiti, D.}, \bibinfo{editor}{Bouchon-Meunier, B.}, \bibinfo{editor}{Yager, R.} (Eds.), \bibinfo{booktitle}{Information Processing and Management of Uncertainty in Knowledge-Based Systems}, \bibinfo{publisher}{Springer}. pp. \bibinfo{pages}{17--30}.
%Type = Inproceedings
\bibitem[{Blocher et~al.(2023)Blocher, Schollmeyer, Jansen and Nalenz}]{blocher23}
\bibinfo{author}{Blocher, H.}, \bibinfo{author}{Schollmeyer, G.}, \bibinfo{author}{Jansen, C.}, \bibinfo{author}{Nalenz, M.}, \bibinfo{year}{2023}.
\newblock \bibinfo{title}{Depth functions for partial orders with a descriptive analysis of machine learning algorithms}, in: \bibinfo{editor}{Miranda, E.}, \bibinfo{editor}{Montes, I.}, \bibinfo{editor}{Quaeghebeur, E.}, \bibinfo{editor}{Vantaggi, B.} (Eds.), \bibinfo{booktitle}{Proceedings of the Thirteenth International Symposium on Imprecise Probability: Theories and Applications}, \bibinfo{publisher}{Proceedings of Machine Learning Research}. pp. \bibinfo{pages}{59--71}.
%Type = Article
\bibitem[{Bradley and Terry(1952)}]{btl}
\bibinfo{author}{Bradley, R.}, \bibinfo{author}{Terry, M.}, \bibinfo{year}{1952}.
\newblock \bibinfo{title}{Rank analysis of incomplete block designs: I. the method of paired comparisons}.
\newblock \bibinfo{journal}{Biometrika} \bibinfo{volume}{39}, \bibinfo{pages}{324--345}.
%Type = Incollection
\bibitem[{Brandenburg et~al.(2012)Brandenburg, Glei{\ss}ner and Hofmeier}]{brandenburg12}
\bibinfo{author}{Brandenburg, F.}, \bibinfo{author}{Glei{\ss}ner, A.}, \bibinfo{author}{Hofmeier, A.}, \bibinfo{year}{2012}.
\newblock \bibinfo{title}{Comparing and aggregating partial orders with {K}endall tau distances.}, in: \bibinfo{editor}{Rahman, S.}, \bibinfo{editor}{Nakano, S.} (Eds.), \bibinfo{booktitle}{WALCOM: Algorithms and Computation 2012}. Lecture Notes in Computer Science, pp. \bibinfo{pages}{88--99}.
%Type = Article
\bibitem[{Cawley and Talbot(2010)}]{JMLR:v11:cawley10a}
\bibinfo{author}{Cawley, G.}, \bibinfo{author}{Talbot, N.}, \bibinfo{year}{2010}.
\newblock \bibinfo{title}{On over-fitting in model selection and subsequent selection bias in performance evaluation}.
\newblock \bibinfo{journal}{Journal of Machine Learning Research} \bibinfo{volume}{11}, \bibinfo{pages}{2079--2107}.
%Type = Inbook
\bibitem[{Chambers and Echenique(2016)}]{chambers16}
\bibinfo{author}{Chambers, C.}, \bibinfo{author}{Echenique, F.}, \bibinfo{year}{2016}.
\newblock \bibinfo{title}{Revealed Preference Theory}. \bibinfo{publisher}{Cambridge University Press}. chapter \bibinfo{chapter}{Stochastic Choice}.
\newblock Econometric Society Monographs, pp. \bibinfo{pages}{95--113}.
%Type = Article
\bibitem[{Chang et~al.(2015)Chang, Jim\'enez-Mart\'in, Maasoumi and P\'erez-Amaral}]{c2015}
\bibinfo{author}{Chang, C.}, \bibinfo{author}{Jim\'enez-Mart\'in, J.}, \bibinfo{author}{Maasoumi, E.}, \bibinfo{author}{P\'erez-Amaral, T.}, \bibinfo{year}{2015}.
\newblock \bibinfo{title}{A stochastic dominance approach to financial risk management strategies}.
\newblock \bibinfo{journal}{Journal of Econometrics} \bibinfo{volume}{187}, \bibinfo{pages}{472--485}.
%Type = Article
\bibitem[{Chang(2020)}]{c2020}
\bibinfo{author}{Chang, L.}, \bibinfo{year}{2020}.
\newblock \bibinfo{title}{Partial order relations for classification comparisons}.
\newblock \bibinfo{journal}{Canadian Journal of Statistics} \bibinfo{volume}{48}, \bibinfo{pages}{152--166}.
%Type = Article
\bibitem[{Couso and Dubois(2014)}]{Couso_2014}
\bibinfo{author}{Couso, I.}, \bibinfo{author}{Dubois, D.}, \bibinfo{year}{2014}.
\newblock \bibinfo{title}{Statistical reasoning with set-valued information: Ontic vs. epistemic views}.
\newblock \bibinfo{journal}{International Journal of Approximate Reasoning} \bibinfo{volume}{55}, \bibinfo{pages}{1502--1518}.
%Type = Book
\bibitem[{Critchlow(1985)}]{critchlow85}
\bibinfo{author}{Critchlow, D.}, \bibinfo{year}{1985}.
\newblock \bibinfo{title}{Metric methods for analyzing partially ranked data}. volume~\bibinfo{volume}{34} of \textit{\bibinfo{series}{Lecture Notes in Statistics}}.
\newblock \bibinfo{publisher}{Springer}.
%Type = Article
\bibitem[{Davidson(1970)}]{davidson70}
\bibinfo{author}{Davidson, R.}, \bibinfo{year}{1970}.
\newblock \bibinfo{title}{On extending the {B}radley-{T}erry model to accommodate ties in paired comparison experiments}.
\newblock \bibinfo{journal}{Journal of the American Statistical Association} \bibinfo{volume}{65}, \bibinfo{pages}{317--328}.
%Type = Article
\bibitem[{Dem\v{s}ar(2006)}]{d2006}
\bibinfo{author}{Dem\v{s}ar, J.}, \bibinfo{year}{2006}.
\newblock \bibinfo{title}{Statistical comparisons of classifiers over multiple data sets}.
\newblock \bibinfo{journal}{Journal of Machine Learning Research} \bibinfo{volume}{7}, \bibinfo{pages}{1--30}.
%Type = Misc
\bibitem[{Dua and Graff(2017)}]{dua17}
\bibinfo{author}{Dua, D.}, \bibinfo{author}{Graff, C.}, \bibinfo{year}{2017}.
\newblock \bibinfo{title}{Uci machine learning repository}.
\newblock \URLprefix \url{http://archive.ics.uci.edu/ml}. \bibinfo{note}{[Accessed: 09.10.2023]}.
%Type = Incollection
\bibitem[{Eckhoff(1993)}]{eckhoff93}
\bibinfo{author}{Eckhoff, J.}, \bibinfo{year}{1993}.
\newblock \bibinfo{title}{Chapter 2.1 - {H}elly, {R}adon, and {C}arath{\'{e}}odory type theorems}, in: \bibinfo{editor}{Gruber, P.}, \bibinfo{editor}{Wwillis, J.} (Eds.), \bibinfo{booktitle}{Handbook of Convex Geometry}. \bibinfo{publisher}{North-Holland}, \bibinfo{address}{Amsterdam}, pp. \bibinfo{pages}{389--448}.
%Type = Article
\bibitem[{Eugster et~al.(2012)Eugster, Hothorn and Leisch}]{ehl2012}
\bibinfo{author}{Eugster, M.}, \bibinfo{author}{Hothorn, T.}, \bibinfo{author}{Leisch, F.}, \bibinfo{year}{2012}.
\newblock \bibinfo{title}{Domain-based benchmark experiments: Exploratory and inferential analysis}.
\newblock \bibinfo{journal}{Austrian Journal of Statistics} \bibinfo{volume}{41}, \bibinfo{pages}{5--26}.
%Type = Article
\bibitem[{Fligner and Verducci(1986)}]{fligner}
\bibinfo{author}{Fligner, M.}, \bibinfo{author}{Verducci, J.}, \bibinfo{year}{1986}.
\newblock \bibinfo{title}{Distance based ranking models}.
\newblock \bibinfo{journal}{Journal of the Royal Statistical Society. Series B (Methodological)} \bibinfo{volume}{48}, \bibinfo{pages}{359--369}.
%Type = Article
\bibitem[{Friedman et~al.(2021)Friedman, Hastie, Tibshirani, Narasimhan, Tay, Simon and Qian}]{friedman2021package}
\bibinfo{author}{Friedman, J.}, \bibinfo{author}{Hastie, T.}, \bibinfo{author}{Tibshirani, R.}, \bibinfo{author}{Narasimhan, B.}, \bibinfo{author}{Tay, K.}, \bibinfo{author}{Simon, N.}, \bibinfo{author}{Qian, J.}, \bibinfo{year}{2021}.
\newblock \bibinfo{title}{Package glmnet}.
\newblock \bibinfo{journal}{CRAN R Repository} .
%Type = Inproceedings
\bibitem[{Ganter(2010)}]{ganter2010two}
\bibinfo{author}{Ganter, B.}, \bibinfo{year}{2010}.
\newblock \bibinfo{title}{Two basic algorithms in concept analysis}, in: \bibinfo{booktitle}{Formal Concept Analysis: 8th International Conference, ICFCA 2010, Agadir, Morocco, March 15-18, 2010. Proceedings 8}, \bibinfo{organization}{Springer}. pp. \bibinfo{pages}{312--340}.
%Type = Book
\bibitem[{Ganter and Wille(2012)}]{ganter12}
\bibinfo{author}{Ganter, B.}, \bibinfo{author}{Wille, R.}, \bibinfo{year}{2012}.
\newblock \bibinfo{title}{Formal Concept Analysis: Mathematical Foundations}.
\newblock \bibinfo{publisher}{Springer}.
%Type = Misc
\bibitem[{Goibert et~al.(2022)Goibert, Cl{'e}men{\c{c}}on, Irurozki and Mozharovskyi}]{goibert22}
\bibinfo{author}{Goibert, M.}, \bibinfo{author}{Cl{'e}men{\c{c}}on, S.}, \bibinfo{author}{Irurozki, E.}, \bibinfo{author}{Mozharovskyi, P.}, \bibinfo{year}{2022}.
\newblock \bibinfo{title}{Statistical depth functions for ranking distributions: Definitions, statistical learning and applications}.
\newblock \URLprefix \url{https://arxiv.org/abs/2201.08105}, \href{http://arxiv.org/abs/2201.08105}{{\tt arXiv:2201.08105}}. \bibinfo{note}{[Accessed: 13.11.2023]}.
%Type = Misc
\bibitem[{Hechenbichler and Schliep(2004)}]{epub1769}
\bibinfo{author}{Hechenbichler, K.}, \bibinfo{author}{Schliep, K.}, \bibinfo{year}{2004}.
\newblock \bibinfo{title}{Weighted k-nearest-neighbor techniques and ordinal classification}.
\newblock \bibinfo{howpublished}{Technical Report, LMU}.
\newblock \URLprefix \url{http://nbn-resolving.de/urn/resolver.pl?urn=nbn:de:bvb:19-epub-1769-9}. \bibinfo{note}{[Accessed: 28.11.2023]}.
%Type = Article
\bibitem[{Hothorn et~al.(2005)Hothorn, Leisch, Zeileis and Hornik}]{Hothorn}
\bibinfo{author}{Hothorn, T.}, \bibinfo{author}{Leisch, F.}, \bibinfo{author}{Zeileis, A.}, \bibinfo{author}{Hornik, K.}, \bibinfo{year}{2005}.
\newblock \bibinfo{title}{The design and analysis of benchmark experiments}.
\newblock \bibinfo{journal}{Journal of Computational and Graphical Statistics} \bibinfo{volume}{14}, \bibinfo{pages}{675--699}.
%Type = Article
\bibitem[{Jansen et~al.(2022)Jansen, Blocher, Augustin and Schollmeyer}]{jbas2022}
\bibinfo{author}{Jansen, C.}, \bibinfo{author}{Blocher, H.}, \bibinfo{author}{Augustin, T.}, \bibinfo{author}{Schollmeyer, G.}, \bibinfo{year}{2022}.
\newblock \bibinfo{title}{Information efficient learning of complexly structured preferences: Elicitation procedures and their application to decision making under uncertainty}.
\newblock \bibinfo{journal}{International Journal of Approximate Reasoning} \bibinfo{volume}{144}, \bibinfo{pages}{69--91}.
%Type = Article
\bibitem[{Jansen et~al.(2023a)Jansen, Nalenz, Schollmeyer and Augustin}]{jansen23}
\bibinfo{author}{Jansen, C.}, \bibinfo{author}{Nalenz, M.}, \bibinfo{author}{Schollmeyer, G.}, \bibinfo{author}{Augustin, T.}, \bibinfo{year}{2023}a.
\newblock \bibinfo{title}{Statistical comparisons of classifiers by generalized stochastic dominance}.
\newblock \bibinfo{journal}{Journal of Machine Learning Research} \bibinfo{volume}{24}, \bibinfo{pages}{1--37}.
%Type = Article
\bibitem[{Jansen et~al.(2018a)Jansen, Schollmeyer and Augustin}]{jsa2018}
\bibinfo{author}{Jansen, C.}, \bibinfo{author}{Schollmeyer, G.}, \bibinfo{author}{Augustin, T.}, \bibinfo{year}{2018}a.
\newblock \bibinfo{title}{Concepts for decision making under severe uncertainty with partial ordinal and partial cardinal preferences}.
\newblock \bibinfo{journal}{International Journal of Approximate Reasoning} \bibinfo{volume}{98}, \bibinfo{pages}{112--131}.
%Type = Article
\bibitem[{Jansen et~al.(2018b)Jansen, Schollmeyer and Augustin}]{jsa2018b}
\bibinfo{author}{Jansen, C.}, \bibinfo{author}{Schollmeyer, G.}, \bibinfo{author}{Augustin, T.}, \bibinfo{year}{2018}b.
\newblock \bibinfo{title}{A probabilistic evaluation framework for preference aggregation reflecting group homogeneity}.
\newblock \bibinfo{journal}{Mathematical Social Sciences} \bibinfo{volume}{96}, \bibinfo{pages}{49--62}.
%Type = Inproceedings
\bibitem[{Jansen et~al.(2023b)Jansen, Schollmeyer and Augustin}]{jsa2023}
\bibinfo{author}{Jansen, C.}, \bibinfo{author}{Schollmeyer, G.}, \bibinfo{author}{Augustin, T.}, \bibinfo{year}{2023}b.
\newblock \bibinfo{title}{Multi-target decision making under conditions of severe uncertainty}, in: \bibinfo{editor}{Torra, V.}, \bibinfo{editor}{Narukawa, Y.} (Eds.), \bibinfo{booktitle}{Modeling Decisions for Artificial Intelligence}, \bibinfo{publisher}{Springer}. pp. \bibinfo{pages}{45--57}.
%Type = Inproceedings
\bibitem[{Jansen et~al.(2023c)Jansen, Schollmeyer, Blocher, Rodemann and Augustin}]{uaiall}
\bibinfo{author}{Jansen, C.}, \bibinfo{author}{Schollmeyer, G.}, \bibinfo{author}{Blocher, H.}, \bibinfo{author}{Rodemann, J.}, \bibinfo{author}{Augustin, T.}, \bibinfo{year}{2023}c.
\newblock \bibinfo{title}{Robust statistical comparison of random variables with locally varying scale of measurement}, in: \bibinfo{editor}{Evans, R.J.}, \bibinfo{editor}{Shpitser, I.} (Eds.), \bibinfo{booktitle}{Proceedings of the Thirty-Ninth Conference on Uncertainty in Artificial Intelligence}, \bibinfo{publisher}{Proceedings of Machine Learning Research}. pp. \bibinfo{pages}{941--952}.
%Type = Article
\bibitem[{Kikuti et~al.(2011)Kikuti, Cozman and Filho}]{kikuti}
\bibinfo{author}{Kikuti, D.}, \bibinfo{author}{Cozman, F.}, \bibinfo{author}{Filho, R.}, \bibinfo{year}{2011}.
\newblock \bibinfo{title}{Sequential decision making with partially ordered preferences}.
\newblock \bibinfo{journal}{Artificial Intelligence} \bibinfo{volume}{175}, \bibinfo{pages}{1346 -- 1365}.
%Type = Article
\bibitem[{Lebanon and Mao(2008)}]{NIPS2007_fe8c15fe}
\bibinfo{author}{Lebanon, G.}, \bibinfo{author}{Mao, Y.}, \bibinfo{year}{2008}.
\newblock \bibinfo{title}{Non-parametric modeling of partially ranked data}.
\newblock \bibinfo{journal}{Journal of Machine Learning Research} \bibinfo{volume}{9}, \bibinfo{pages}{2401--2429}.
%Type = Article
\bibitem[{Levy and Levy(1984)}]{ll1984}
\bibinfo{author}{Levy, H.}, \bibinfo{author}{Levy, A.}, \bibinfo{year}{1984}.
\newblock \bibinfo{title}{Ordering uncertain options under inflation: A note}.
\newblock \bibinfo{journal}{The Journal of Finance} \bibinfo{volume}{39}, \bibinfo{pages}{1223--1229}.
%Type = Article
\bibitem[{Liu(1990)}]{liu90}
\bibinfo{author}{Liu, R.}, \bibinfo{year}{1990}.
\newblock \bibinfo{title}{On a notion of data depth based on random simplices}.
\newblock \bibinfo{journal}{The Annals of Statistics} \bibinfo{volume}{18}, \bibinfo{pages}{405--414}.
%Type = Inproceedings
\bibitem[{Mau\'a et~al.(2017)Mau\'a, Cozman, Conaty and Campos}]{pmlr-v62-maua17a}
\bibinfo{author}{Mau\'a, D.}, \bibinfo{author}{Cozman, F.}, \bibinfo{author}{Conaty, D.}, \bibinfo{author}{Campos, C.}, \bibinfo{year}{2017}.
\newblock \bibinfo{title}{Credal sum-product networks}, in: \bibinfo{editor}{Antonucci, A.}, \bibinfo{editor}{Corani, G.}, \bibinfo{editor}{Couso, I.}, \bibinfo{editor}{Destercke, S.} (Eds.), \bibinfo{booktitle}{Proceedings of the Tenth International Symposium on Imprecise Probability: Theories and Applications}, \bibinfo{publisher}{Proceedings of Machine Learning Research}. pp. \bibinfo{pages}{205--216}.
%Type = Book
\bibitem[{Mosler(2002)}]{m2002}
\bibinfo{author}{Mosler, K.}, \bibinfo{year}{2002}.
\newblock \bibinfo{title}{Multivariate Dispersion, Central Regions, and Depth: The Lift Zonoid Approach}.
\newblock \bibinfo{publisher}{Springer}.
%Type = Article
\bibitem[{Mosler and Mozharovskyi(2022)}]{mosler22}
\bibinfo{author}{Mosler, K.}, \bibinfo{author}{Mozharovskyi, P.}, \bibinfo{year}{2022}.
\newblock \bibinfo{title}{Choosing among notions of multivariate depth statistics}.
\newblock \bibinfo{journal}{Statistical Science} \bibinfo{volume}{37}, \bibinfo{pages}{348--368}.
%Type = Misc
\bibitem[{Nakamura et~al.(2019)Nakamura, Yano and Komaki}]{nakamura2019learning}
\bibinfo{author}{Nakamura, K.}, \bibinfo{author}{Yano, K.}, \bibinfo{author}{Komaki, F.}, \bibinfo{year}{2019}.
\newblock \bibinfo{title}{Learning partially ranked data based on graph regularization}.
\newblock \href{http://arxiv.org/abs/1902.10963}{{\tt arXiv:1902.10963}}. \bibinfo{note}{[accessed: 28.11.2023]}.
%Type = Article
\bibitem[{Pini et~al.(2011)Pini, Rossi, Venable and Walsh}]{p2012}
\bibinfo{author}{Pini, M.}, \bibinfo{author}{Rossi, F.}, \bibinfo{author}{Venable, K.}, \bibinfo{author}{Walsh, T.}, \bibinfo{year}{2011}.
\newblock \bibinfo{title}{Incompleteness and incomparability in preference aggregation: Complexity results}.
\newblock \bibinfo{journal}{Artificial Intelligence} \bibinfo{volume}{175}, \bibinfo{pages}{1272--1289}.
%Type = Article
\bibitem[{Plackett(1975)}]{plackett75}
\bibinfo{author}{Plackett, R.}, \bibinfo{year}{1975}.
\newblock \bibinfo{title}{The analysis of permutations}.
\newblock \bibinfo{journal}{Journal of the Royal Statistical Society Series C: Applied Statistics} \bibinfo{volume}{24}, \bibinfo{pages}{193--202}.
%Type = Inproceedings
\bibitem[{Plass et~al.(2015a)Plass, Augustin, Cattaneo and Schollmeyer}]{plass2015statistical}
\bibinfo{author}{Plass, J.}, \bibinfo{author}{Augustin, T.}, \bibinfo{author}{Cattaneo, M.}, \bibinfo{author}{Schollmeyer, G.}, \bibinfo{year}{2015}a.
\newblock \bibinfo{title}{Statistical modelling under epistemic data imprecision: some results on estimating multinomial distributions and logistic regression for coarse categorical data}, in: \bibinfo{editor}{Augustin, T.}, \bibinfo{editor}{Doria, S.}, \bibinfo{editor}{Miranda, E.}, \bibinfo{editor}{Quaeghebeur, E.} (Eds.), \bibinfo{booktitle}{Proceedings of the Ninth International Symposium on Imprecise Probability: Theories and Applications}, \bibinfo{publisher}{Aracne}. pp. \bibinfo{pages}{247--256}.
%Type = Inproceedings
\bibitem[{Plass et~al.(2015b)Plass, Fink, Sch\"oning and Augustin}]{plass2015_ontic}
\bibinfo{author}{Plass, J.}, \bibinfo{author}{Fink, P.}, \bibinfo{author}{Sch\"oning, N.}, \bibinfo{author}{Augustin, T.}, \bibinfo{year}{2015}b.
\newblock \bibinfo{title}{Statistical modelling in surveys without neglecting the undecided: Multinomial logistic regression models and imprecise classification trees under ontic data imprecision}, in: \bibinfo{editor}{Augustin, T.}, \bibinfo{editor}{Doria, S.}, \bibinfo{editor}{Miranda, E.}, \bibinfo{editor}{Quaeghebeur, E.} (Eds.), \bibinfo{booktitle}{Proceedings of the Ninth International Symposium on Imprecise Probability: Theories and Applications}, \bibinfo{publisher}{Aracne}. pp. \bibinfo{pages}{257--266}.
%Type = Article
\bibitem[{Rao and Kupper(1967)}]{rao67}
\bibinfo{author}{Rao, P.V.}, \bibinfo{author}{Kupper, L.}, \bibinfo{year}{1967}.
\newblock \bibinfo{title}{Ties in paired-comparison experiments: A generalization of the bradley-terry model}.
\newblock \bibinfo{journal}{Journal of the American Statistical Association} \bibinfo{volume}{62}, \bibinfo{pages}{194--204}.
%Type = Misc
\bibitem[{Schollmeyer(2017a)}]{schollmeyer17b}
\bibinfo{author}{Schollmeyer, G.}, \bibinfo{year}{2017}a.
\newblock \bibinfo{title}{Application of lower quantiles for complete lattices to ranking data: Analyzing outlyingness of preference orderings}.
\newblock \bibinfo{howpublished}{Technical Report, LMU}.
\newblock \URLprefix \url{http://nbn-resolving.de/urn/resolver.pl?urn=nbn:de:bvb:19-epub-40452-9}. \bibinfo{note}{[Accessed: 28.11.2023]}.
%Type = Misc
\bibitem[{Schollmeyer(2017b)}]{schollmeyer17a}
\bibinfo{author}{Schollmeyer, G.}, \bibinfo{year}{2017}b.
\newblock \bibinfo{title}{Lower quantiles for complete lattices}.
\newblock \bibinfo{howpublished}{Technical Report, LMU}.
\newblock \URLprefix \url{http://nbn-resolving.de/urn/resolver.pl?urn=nbn:de:bvb:19-epub-40448-7}. \bibinfo{note}{[Accessed: 28.11.2023]}.
%Type = Inproceedings
\bibitem[{Schollmeyer(2019)}]{pmlr-v103-schollmeyer19a}
\bibinfo{author}{Schollmeyer, G.}, \bibinfo{year}{2019}.
\newblock \bibinfo{title}{A short note on the equivalence of the ontic and the epistemic view on data imprecision for the case of stochastic dominance for interval-valued data}, in: \bibinfo{editor}{De~Bock, J.}, \bibinfo{editor}{de~Campos, C.}, \bibinfo{editor}{de~Cooman, G.}, \bibinfo{editor}{Quaeghebeur, E.}, \bibinfo{editor}{Wheeler, G.} (Eds.), \bibinfo{booktitle}{Proceedings of the Eleventh International Symposium on Imprecise Probabilities: Theories and Applications}, \bibinfo{publisher}{Proceedings of Machine Learning Research}. pp. \bibinfo{pages}{330--337}.
%Type = Misc
\bibitem[{Schollmeyer et~al.(2017)Schollmeyer, Jansen and Augustin}]{epub40416}
\bibinfo{author}{Schollmeyer, G.}, \bibinfo{author}{Jansen, C.}, \bibinfo{author}{Augustin, T.}, \bibinfo{year}{2017}.
\newblock \bibinfo{title}{Detecting stochastic dominance for poset-valued random variables as an example of linear programming on closure systems}.
\newblock \bibinfo{howpublished}{Technical Report, LMU}.
\newblock \URLprefix \url{http://nbn-resolving.de/urn/resolver.pl?urn=nbn:de:bvb:19-epub-40416-0}. \bibinfo{note}{[Accessed: 28.11.2023]}.
%Type = Article
\bibitem[{Seidenfeld et~al.(1995)Seidenfeld, Kadane and Schervish}]{sks1995}
\bibinfo{author}{Seidenfeld, T.}, \bibinfo{author}{Kadane, J.}, \bibinfo{author}{Schervish, M.}, \bibinfo{year}{1995}.
\newblock \bibinfo{title}{A representation of partially ordered preferences}.
\newblock \bibinfo{journal}{Annals of Statistics} \bibinfo{volume}{23}, \bibinfo{pages}{2168--2217}.
%Type = Incollection
\bibitem[{Sinclair(1982)}]{sinclair82}
\bibinfo{author}{Sinclair, C.D.}, \bibinfo{year}{1982}.
\newblock \bibinfo{title}{Glim for preference}, in: \bibinfo{editor}{Gilchrist, R.} (Ed.), \bibinfo{booktitle}{GLIM 82: Proceedings of the International Conference on Generalised Linear Models}. \bibinfo{publisher}{Springer}, pp. \bibinfo{pages}{164--178}.
%Type = Inproceedings
\bibitem[{Stoye(2009)}]{stoye2009statistical}
\bibinfo{author}{Stoye, J.}, \bibinfo{year}{2009}.
\newblock \bibinfo{title}{Statistical inference for interval identified parameters}, in: \bibinfo{editor}{Augustin, T.}, \bibinfo{editor}{Coolen, F.}, \bibinfo{editor}{Moral, S.}, \bibinfo{editor}{Troffaes, M.} (Eds.), \bibinfo{booktitle}{Proceedings of the Sixth International Symposium on Imprecise Probabilities: Theories and Applications}, \bibinfo{publisher}{Aracne}. pp. \bibinfo{pages}{395--404}.
%Type = Misc
\bibitem[{Therneau et~al.(2015)Therneau, Atkinson and Ripley}]{therneau2015package}
\bibinfo{author}{Therneau, T.}, \bibinfo{author}{Atkinson, B.}, \bibinfo{author}{Ripley, B.}, \bibinfo{year}{2015}.
\newblock \bibinfo{title}{Package rpart}.
\newblock \URLprefix \url{http://cran.ma.ic.ac.uk/web/packages/rpart/rpart.pdf}. \bibinfo{note}{[Accessed: 15.02.2023]}.
%Type = Article
\bibitem[{Trotter(1974)}]{TROTTER197485}
\bibinfo{author}{Trotter, W.}, \bibinfo{year}{1974}.
\newblock \bibinfo{title}{Dimension of the crown skn}.
\newblock \bibinfo{journal}{Discrete Mathematics} \bibinfo{volume}{8}, \bibinfo{pages}{85--103}.
%Type = Inproceedings
\bibitem[{Tukey(1975)}]{tukey75}
\bibinfo{author}{Tukey, J.}, \bibinfo{year}{1975}.
\newblock \bibinfo{title}{Mathematics and the picturing of data}, in: \bibinfo{editor}{James, R.} (Ed.), \bibinfo{booktitle}{Proceedings of the International Congress of Mathematicians Vancouver}, \bibinfo{publisher}{Mathematics-Congresses}, \bibinfo{address}{Vancouver}. pp. \bibinfo{pages}{523--531}.
%Type = Article
\bibitem[{Vanschoren et~al.(2013)Vanschoren, van Rijn, Bischl and Torgo}]{OpenML2013}
\bibinfo{author}{Vanschoren, J.}, \bibinfo{author}{van Rijn, J.}, \bibinfo{author}{Bischl, B.}, \bibinfo{author}{Torgo, L.}, \bibinfo{year}{2013}.
\newblock \bibinfo{title}{Openml: Networked science in machine learning}.
\newblock \bibinfo{journal}{SIGKDD Explorations} \bibinfo{volume}{15}, \bibinfo{pages}{49--60}.
%Type = Inproceedings
\bibitem[{Vapnik and Chervonenkis(2015)}]{Vapnik15}
\bibinfo{author}{Vapnik, V.}, \bibinfo{author}{Chervonenkis, A.}, \bibinfo{year}{2015}.
\newblock \bibinfo{title}{On the uniform convergence of relative frequencies of events to their probabilities}, in: \bibinfo{editor}{Vovk, V.}, \bibinfo{editor}{Papadopoulos, H.}, \bibinfo{editor}{Gammerman, A.} (Eds.), \bibinfo{booktitle}{Measures of Complexity: Festschrift for Alexey Chervonenkis}, \bibinfo{publisher}{Springer}. pp. \bibinfo{pages}{11--30}.
%Type = Article
\bibitem[{Varma and Simon(2006)}]{varma2006bias}
\bibinfo{author}{Varma, S.}, \bibinfo{author}{Simon, R.}, \bibinfo{year}{2006}.
\newblock \bibinfo{title}{Bias in error estimation when using cross-validation for model selection}.
\newblock \bibinfo{journal}{BMC Bioinformatics} \bibinfo{volume}{7}, \bibinfo{pages}{1--8}.
%Type = Article
\bibitem[{Wright and Ziegler(2017)}]{ranger}
\bibinfo{author}{Wright, M.}, \bibinfo{author}{Ziegler, A.}, \bibinfo{year}{2017}.
\newblock \bibinfo{title}{{ranger}: A fast implementation of random forests for high dimensional data in {C++} and {R}}.
\newblock \bibinfo{journal}{Journal of Statistical Software} \bibinfo{volume}{77}, \bibinfo{pages}{1--17}.
%Type = Article
\bibitem[{Zaffalon(2002)}]{ZAFFALON20025}
\bibinfo{author}{Zaffalon, M.}, \bibinfo{year}{2002}.
\newblock \bibinfo{title}{The naive credal classifier}.
\newblock \bibinfo{journal}{Journal of Statistical Planning and Inference} \bibinfo{volume}{105}, \bibinfo{pages}{5--21}.
%Type = Article
\bibitem[{Zaffalon et~al.(2012)Zaffalon, Corani and Mau\'a}]{ZAFFALON20121282}
\bibinfo{author}{Zaffalon, M.}, \bibinfo{author}{Corani, G.}, \bibinfo{author}{Mau\'a, D.}, \bibinfo{year}{2012}.
\newblock \bibinfo{title}{Evaluating credal classifiers by utility-discounted predictive accuracy}.
\newblock \bibinfo{journal}{International Journal of Approximate Reasoning} \bibinfo{volume}{53}, \bibinfo{pages}{1282--1301}.
%Type = Article
\bibitem[{Zuo and Serfling(2000)}]{sz2000}
\bibinfo{author}{Zuo, Y.}, \bibinfo{author}{Serfling, R.}, \bibinfo{year}{2000}.
\newblock \bibinfo{title}{{General notions of statistical depth function}}.
\newblock \bibinfo{journal}{The Annals of Statistics} \bibinfo{volume}{28}, \bibinfo{pages}{461 -- 482}.

\end{thebibliography}

% TeX file.

%\begin{thebibliography}{00}
%
%%% \bibitem[Author(year)]{label}
%%% Text of bibliographic item
%\end{thebibliography}
\end{document}